\documentclass[10pt,twocolumn,letterpaper]{article}

\usepackage{iccv}
\usepackage{times}
\usepackage{epsfig}
\usepackage{graphicx}
\usepackage{mwe}
\usepackage{amsmath}
\usepackage{amssymb}
\usepackage{pifont}
\usepackage{bbm}
\usepackage{subcaption}
\usepackage{multirow}
\usepackage{floatrow}
\usepackage{float}
\usepackage{color, colortbl}
\usepackage{xcolor}
\usepackage{booktabs}
\usepackage[vlined,ruled]{algorithm2e} 
\usepackage{algpseudocode}
\usepackage{diagbox}
\usepackage[toc,page,header]{appendix}
\usepackage{minitoc}

\newfloatcommand{capbtabbox}{table}
\newfloatcommand{capbfigbox}{figure}

\DeclareMathOperator*{\argmin}{argmin}
\DeclareMathOperator*{\argmax}{argmax}
\DeclareMathOperator{\Var}{Var}
\DeclareMathOperator*{\md}{\dot{\implies}}

\newcommand{\sone}{\texttt{AmCo}\xspace}%
\newcommand{\stwo}{\texttt{CoSi}\xspace}%
\definecolor{Gray}{gray}{0.9}
\newcommand{\nhi}{NICO$++^{95}$}
\newcommand{\nmi}{NICO$++^{90}$}
\newcommand{\nlo}{NICO$++^{75}$}
\newcommand{\pak}{\texttt{Precision@k}\xspace}
\newcommand{\paten}{\texttt{Precision@10}\xspace}
\newcommand{\cohterm}
{\textit{coherence prior}\xspace}
\newcommand{\predterm}
{\textit{correlation prior}\xspace}

\newcommand{\methodname}{First Amplify Correlations and Then Slice\xspace}
\newcommand{\method}{\texttt{FACTS}\xspace} %
\newcommand{\varconf}{$\sigma_\sone$\xspace}
\newcommand{\varconfeq}{\sigma_\texttt{AmCo}}
\newcommand{\avgap}{\texttt{Avg-AP}\xspace}

\usepackage[pagebackref=true,breaklinks=true,letterpaper=true,colorlinks,bookmarks=false]{hyperref}
\usepackage[accsupp]{axessibility}
 \iccvfinalcopy %

\def\iccvPaperID{1033} %

\ificcvfinal\fi
\begin{document}
\definecolor{lightgray}{RGB}{220,220,220}
\doparttoc %
\faketableofcontents
\newcommand{\jh}[1]{\textcolor{magenta}{Judy: #1}}

\title{FACTS: First Amplify Correlations and Then Slice to Discover Bias}

\author{%
    Sriram Yenamandra
    \quad
    Pratik Ramesh
    \quad
    Viraj Prabhu  
    \quad
      Judy Hoffman \\
Georgia Institute of Technology \\
{\tt\small \{sriramy, pratikramesh, virajp, judy\}@gatech.edu}}

\maketitle
\ificcvfinal\thispagestyle{empty}\fi
\everypar{\looseness=-1}

\begin{abstract}
\vspace{-5pt}
Computer vision datasets frequently contain spurious correlations between task-relevant labels and (easy to learn) latent task-irrelevant attributes (\eg context). Models trained on such datasets learn “shortcuts” and underperform on bias-conflicting slices of data where the correlation does not hold. In this work, we study the problem of identifying such slices to inform downstream bias mitigation strategies. We propose \methodname (\method), wherein we first amplify correlations to fit a simple bias-aligned hypothesis via strongly regularized empirical risk minimization. Next, we perform correlation-aware slicing via mixture modeling in bias-aligned feature space to discover underperforming data slices that capture distinct correlations. Despite its simplicity, our method considerably improves over prior work (by as much as 35\% precision@10) in correlation bias identification across a range of diverse evaluation settings. Code:  \url{https://github.com/yvsriram/FACTS}.
\vspace{-5pt}
\end{abstract}

\section{Introduction}

\begin{figure}[t]
\includegraphics[width=\linewidth]{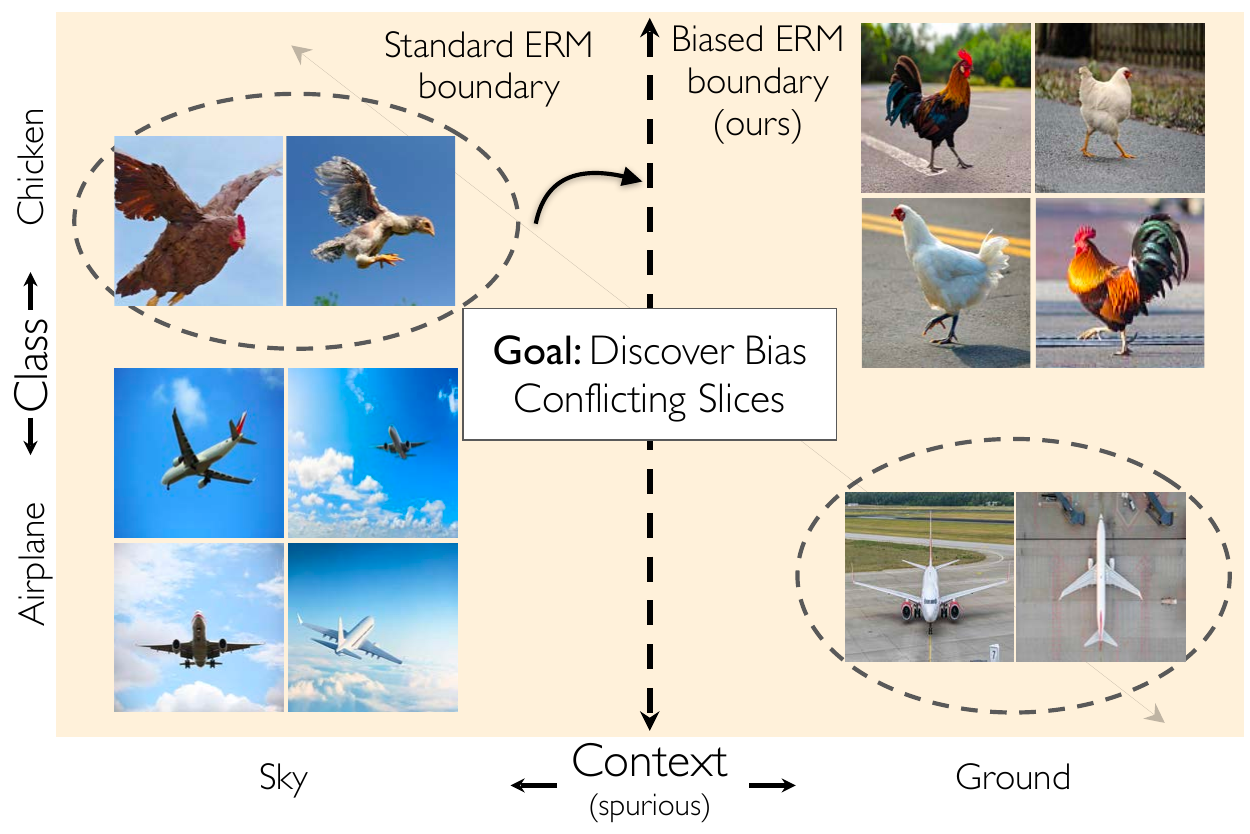}
\vspace{-15pt}
\caption{Consider the task of classifying images of \emph{chickens} and \emph{airplanes}. Real-world datasets typically also contain spurious correlations between (easy to learn) task-irrelevant attributes and labels: \eg most \emph{airplane} images are airborne, whereas \emph{chickens} are usually photographed on the ground. Naively training on such a dataset would lead to overfitting to the majority context within each class, and to underperforming on bias-\emph{conflicting} slices of data (\eg \emph{chickens} in the \emph{sky}). We study the problem of automatically identifying such bias-conflicting slices (dashed ovals), which can then inform downstream mitigation strategies. Our method, \method, first \emph{amplifies correlations} to learn a context-aligned decision boundary, and then \emph{clusters} samples that are confidently misclassified by this biased model to uncover bias-conflicting slices.}
\label{fig:teaser_fig}
\end{figure}

Real-world datasets frequently exhibit \textit{correlation biases}, wherein a task-\emph{irrelevant} attribute (say, the image background) is correlated with the task label of interest~\cite{torralba2011unbiased,caliskan2017semantics,buolamwini2018gender}. 
Consider the task of distinguishing images of \emph{chickens} from \emph{airplanes} (see Fig.~\ref{fig:teaser_fig}). Naturally, most images of \textit{airplanes} are in the \textit{sky}, whereas \textit{chickens} are typically found on the \textit{ground}. 
A naive classifier trained on this biased dataset may conceivably learn to over-rely on the (relatively easier to learn) background and consequently underperform on \emph{bias-conflicting} slices of the data (\eg \textit{chickens} in the \textit{air}). Indeed, deep models trained with standard empirical risk minimization~\cite{vapnik1999nature} are notorious for exploiting such ``shortcuts''~\cite{geirhos2020shortcut, jabbour2020-pmlr-v126}, which may have serious repercussions in high-stakes applications like medicine~\cite{badgeley2019deep,winkler2019association}, face recognition~\cite{buolamwini2018gender}, and autonomous driving~\cite{wilson2019predictive}.  

In this work, given a potentially biased dataset, we attempt to \emph{automatically} identify such bias-conflicting slices: dataset subsets wherein the spurious correlation \emph{does not hold}. 
Such identification could inform downstream mitigation strategies based on reweighting ~\cite{nam2022spread, sagawa2020distributionally, liu2021jtt, nam2020learning, li2023bias} or annotating \cite{jain2022distilling, chawla2002smote, he2008adasyn} more instances of underrepresented populations. Importantly, such a method should be able to discover slices that represent \emph{semantically coherent} and \emph{distinct} bias-conflicting subpopulations, that may optionally be ``named'' (say, using an image captioning model~\cite{mokady2021clipcap}), and presented to a practitioner. 

Why is discovering bias-conflicting slices hard? Presumably, one could annotate and control for potentially spurious attributes~\cite{wong2021leveraging,singla2022salient,shah2022modeldiff}, but this is challenging to scale to larger datasets. Further, tasks labels are sometimes spuriously correlated with \emph{latent} \cite{jabbour2020-pmlr-v126} attributes which may be unknown apriori \cite{oakden2020meaningfulfailures}, or not correspond to a clean, ``human-interpretable'' concept~\cite{krishnakumar2021udis} to begin with!

Some recent works have focused on fully automated solutions to this problem by posing it as a problem of discovering \emph{systematic error modes} on a held out validation set (via ``error-aware'' mixture modeling~\cite{eyuboglu2022domino} or distilling model failures as directions in latent space ~\cite{jain2022distilling}), or by using external pretrained models such as CLIP~\cite{zhang2023diagnosing} for image captioning~\cite{kim2023explaining}. However, 
we find that these methods can either only diagnose a single bias-conflicting slice per-class~\cite{jain2022distilling}, fail to generalize to settings with severe correlation bias~\cite{eyuboglu2022domino}, or do not control for the bias in the external pretrained model~\cite{zhang2023diagnosing,kim2023explaining} itself. Further, by focusing exclusively on failure modes, such methods may miss underrepresented subpopulations which an overparameterized model may have memorized but still does not understand~\cite{bansal2022measures}.

To address these limitations, we propose a simple algorithm we call \methodname (\method). Our method first amplifies the model's reliance on the underlying spurious correlation by training with heavily regularized empirical risk minimization. By doing so, we force the model to fit a simple, bias-aligned hypothesis that maximally separates bias-conflicting and bias-aligned samples within each class, making them easier to segregate. Next, we propose a novel slicing strategy, that fits per-class mixture models in bias-amplified feature space (to ensure correlation-aware clustering) with an additional coherence prior (to ensure semantic coherence). By making limited assumptions, our method is able to generalize to challenging but practical evaluation settings not previously considered in the literature~\cite{eyuboglu2022domino,jain2022distilling}:
containing \emph{multiple} minority groups per-class, 
or containing a class without a minority group. We make the following contributions:
\begin{itemize}
    \item We study the problem of automatically discovering coherent and distinct slices of a dataset containing correlation bias where the correlation does not hold.
    \item We propose \method, a novel two-stage algorithm that clusters in correlation-amplified feature space to uncover bias-conflicting slices, without assuming access to additional annotations.
    \item We report results for slice discovery on a range of diverse evaluation settings constructed from the WaterBirds~\cite{sagawa2020distributionally}, CelebA~\cite{liu2015faceattributes}, and the newly introduced NICO++~\cite{zhang2022nico} datasets, and demonstrate strong gains over prior work (with absolute gains of as much as 35\% precision points across datasets!).
\end{itemize}
\section{Related Work}

\noindent\textbf{Discovering Error Modes with Human Supervision.} A number of prior works~\cite{wong2021leveraging, singla2021understanding, shah2022modeldiff} propose human-in-the-loop approaches to characterize model failure modes. Wong~\emph{et al.}~\cite{wong2021leveraging} fits sparse linear layers over deep representations and asks human annotators to verify if the learned features are spurious. Singla~\emph{et al.}~\cite{singla2022salient} have humans annotate spurious neural features from an adversarially robust model for a few highly-activated images, that are used to automatically annotate the remaining dataset. 
While effective, these methods require human supervision or are designed for restricted model classes \emph{e.g.} adversarially robust models. In contrast, we propose a fully automated approach for discovering correlation bias with minimal assumptions.

\vspace{5pt}
\noindent\textbf{Automated Error Mode discovery.} Recent works have attempted unsupervised discovery of model failure modes.
Singla~\emph{et al.}~\cite{singla2021understanding} learn decision trees over feature representations of misclassified instances from an adversarially robust model. Domino~\cite{eyuboglu2022domino} performs ``error-aware'' clustering of cross-modal embeddings to discover error modes. Jain~\emph{et al.}~\cite{jain2022distilling} distill model failures as directions in latent space by learning an SVM classifier to identify consistent error patterns. DrML~\cite{zhang2023diagnosing} learns a task model on top of a multimodal embedding space, and uses text embeddings to probe the model and identify visual error modes. 
However, these works either assume knowledge apriori of an exhaustive set of possible spurious attributes, or do not generalize to more challenging use-cases (such as multiple bias conflicting attributes per-class). In this work, we focus on the related problem of discovering bias-conflicting slices of a dataset wherein a spurious correlation does not hold. We propose a novel approach that does not assume prior knowledge of potential correlations and generalizes to a diverse range of practical discovery scenarios.

\begin{figure*}[t]
    \centering    \includegraphics[width=\textwidth]{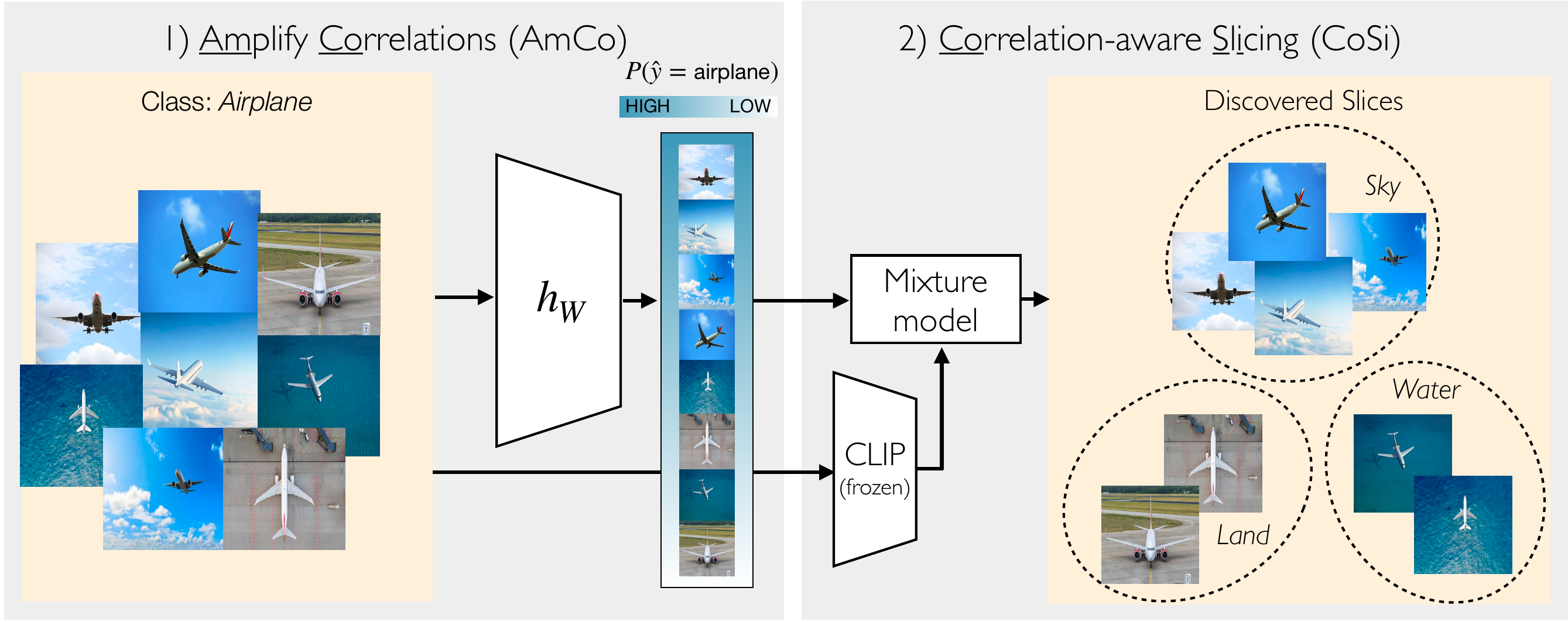}
    \caption{\textbf{\methodname (\method).}
     We seek to identify \emph{bias-conflicting} slices of data where a spurious correlation between a task-irrelevant attribute (\eg background) and the task label of interest \emph{does not hold}. In the example above, this corresponds to \textit{airplane} in \textit{water} or \textit{land}, while  \textit{airplane} in the \textit{sky} forms a bias-aligned slice.
     \textbf{Stage 1}: We first Amplify Correlations (\sone), wherein we learn a simple bias-aligned hypothesis that maximally separates bias-aligned and bias-conflicting samples within each class. \textbf{Stage 2}: Next, we perform Correlation-aware Slicing (\stwo), wherein we perform clustering in bias-amplified feature space, using additional  
     cross-modal CLIP~\cite{radford2021learning} embeddings as a semantic coherence prior. Finally, we present the top-k samples from each discovered slice to a practitioner. }
    \label{fig:approach_fig}
\end{figure*}

\vspace{5pt}
\noindent\textbf{Error Mode Discovery for Bias Mitigation.} A few works discover error modes as an intermediate step toward bias mitigation. 
JTT~\cite{liu2021jtt} learns a model that upweights examples misclassified by a standard ERM model. LfF~\cite{nam2020learning} first emphasizes learning easy samples using a generalized cross-entropy objective, in conjunction with a debiased model that upweights bias-conflicting datapoints identified by the biased model.  Recently, AGRO~\cite{paranjape2023agro} uses  an adversarial slicing model for identifying a group assignment that maximizes the worst-group loss. BAM~\cite{li2023bias} first amplifies bias via introducing auxiliary variables and using a squared loss and then learns a debiased model that upweights samples misclassified by the bias-amplified model. Finally, MAPLE~\cite{zhou22d-pmlr-v162} implements a model agnostic, bi-level formulation for effective sample re-weighting to improve out-of-distribution performance. Our work also seeks to emphasize the model's reliance on spurious correlations to facilitate the discovery of correlation slices by encouraging a bigger separation between bias-aligned and bias-conflicting samples. Unlike prior works that simply upweight underrepresented subpopulations, we
propose a novel algorithm based on bias-amplified clustering that can discover coherent failure modes in diverse evaluation settings.

\section{Method}

\noindent\textbf{Problem Formulation.}
Let $\mathcal{X}$ and $\mathcal{Y}$ denote input and output spaces. Consider a classification task defined over a labeled dataset $D$ with samples $(x, y) \in D$ drawn from $\mathcal{X} \times \mathcal{Y}$. Let $A$ be the set of all spurious attributes across $\mathcal{X}$, where $A = \{a_1,...,a_m\}$.
Let $a_i(x) \in \{0, 1\}$ indicate the presence of an attribute $a_i$ (eg. \emph{sky}, or \emph{road}) for sample $x$. 
We consider an attribute $a$ as \textit{spurious} if (i) it is easier to learn than the target label $y$ (ii) its presence results in the label predominantly assuming a particular value $\hat{y}$, \ie $a$ \textit{mostly dictates} $\hat{y}$, that we denote as $a \md \hat{y}$. Let $M$ be a mapping $M: A \rightarrow \mathcal{Y}$ which matches each spurious attribute to the label it \textit{mostly dictates}. Formally, $\forall a_i \in A$ there exists a unique $M(a_i) \in \mathcal Y$ such that: 
\begin{align}
 \frac{\sum_{(x, y)\in D}{\mathbbm{1}[a_i(x)=1 \text{ and } y=M(a_i)]}}{\sum_{x\in D}a_i(x)} \geq \beta 
\end{align}
where $\beta$ is very large (in our experiments,  typically $\geq 0.7$).

We consider dataset $D$ as containing correlation bias if there exists at least one such spurious attribute. We further assume that all samples have at most one spurious attribute. For a given attribute-label pair $(a_i, y^*)$, let $s(a_i, y^*)$ denote a dataset \textit{slice} satisfying it: $s(a_i, y^*) = \{x \mid \forall (x, y) \in D: y = y^* \text{ and } a_i(x) = 1\}$. Assuming that the set of labels, $\mathcal{Y}$ is known and the set of attributes, $\mathcal{A}$ is unknown. Our goal is to identify the set of \textit{bias-conflicting} slices $S$:
\begin{align*}
    S = \{ s(a, y) \mid  \forall y \in \mathcal Y, \forall a \in A: M(a) \neq y\}
\end{align*}

We now introduce our method, \textbf{F}irst \textbf{A}mplify \textbf{C}orrelations \textbf{T}hen \textbf{S}lice (\method), a two-stage algorithm that automatically identifies coherent bias-conflicting slices of the data by first Amplifying Correlations (\sone), followed by Correlation-aware Slicing (\stwo).

\subsection{First: \underline{Am}plify \underline{Co}rrelations (\sone)}
\label{sec:bai}

Consider dataset $D$ with a spurious correlation $a \md y$. Let $h_{W}$ denote a model parameterized by $W$ trained on this dataset with empirical risk minimization (ERM) \cite{vapnik1999nature}.
Normally, $h_{W}$ will learn a ``shortcut''\cite{geirhos2020shortcut} for predicting $y$ largely based on the spurious attribute $a$. Intuitively, a perfectly bias-aligned decision boundary that solely utilizes $y$ to predict $a$, would yield a feature space in which within-class bias-aligned and bias-conflicting samples are \emph{maximally separated} (\eg \emph{chicken} in the \emph{sky} v/s the \emph{air} in Fig.~\ref{fig:teaser_fig}). Naturally, such a feature space will be highly conducive to segragating bias-conflicting and bias-aligned slices.

In practice however, $h_{W}$ may have a high model capacity and learn both task-relevant and irrelevant features, rather than learning a perfectly bias-aligned decision boundary \cite{hall2022bias}. To overcome this, we propose to \emph{amplify} the model's reliance on the spurious attribute $a$ by restricting its capacity, forcing it to learn a simple hypothesis. In practice, we achieve this by setting the weight decay rate $\lambda$ to a large value during training. We amplify correlations and learn a model $h_{W}$ by minimizing:
 \begin{equation}
    \label{eq:L2_loss}
    \argmin_{W} \quad  \mathbb{E}_{(x, y) \stackrel{\text{bal}}{\in} \mathcal{D}} \mathcal{L}_{CE}(h_W(x), y) + \lambda \lVert W \rVert_2,
\end{equation}
 \noindent where $L_{CE}$ represents a cross-entropy loss and $\text{bal}$ denotes class-balanced sampling, which is required to prevent predictive mode collapse to a single class (the model learning to only predict the
majority class).
  
\noindent\textbf{Restricting model capacity by $\lambda$.} Recall that our goal in this first phase is to learn a model that fits only bias-aligned samples with high confidence and confidently mispredicts bias-conflicting samples. 
 To find a value of $\lambda$ that will result in fitting such a model, we run a hyperparameter sweep across a range of values and for each value retrieve a checkpoint $h_{W^*(\lambda)}$ corresponding to the point at which training accuracy peaks. We limit our search to maximum training accuracy to be able to fairly compare models with varying model capacity by ensuring that each model has fit its training data `optimally'. 
 For a given sample $(x, y) \in D$, let the likelihood of the correct label $y$ under a model $h_W$ be:
   \begin{align*}
     \mathfrak{L}(h_W, x, y) = \texttt{softmax}(h_W(x))[y]
 \end{align*} 
 We compute \varconf($\lambda$), which measures the average per-class \emph{variance} in $\mathfrak{L}(h_{W^*(\lambda)}, x, y)$. Let $D_c$ denote the subset of data samples belonging to class $c \in \mathcal{Y}$. We compute:
 \begin{align}
     \varconfeq(\lambda) = \frac{1}{|\mathcal{Y}|} \sum_{c=1}^{|\mathcal{Y}|} \underset{(x, y){\in}D_c}{\Var}\left[\mathfrak{L}(h_{W^*(\lambda)}, x, y)\right]
 \end{align}

We pick a value $\lambda{=}\lambda^*$ that maximizes \varconf($\lambda$), and use the corresponding model checkpoint $h_{\sone}=h_{W^*(\lambda^*)}$
for the next phase of our method. Intuitively, \varconf captures the separation between features for bias-aligned and bias-conflicting samples within a class: a high value indicates large separation, which indicates that the model has learned a heavily bias-aligned decision boundary (see Fig.~\ref{fig:teaser_fig}, and empirical verification in Sec.~\ref{subsec:analysis}), a property we explicitly leverage in the next stage of our method. 
 
 Fig.~\ref{fig:approach_fig} (\emph{left}) illustrates the \sone phase, wherein we learn a simple bias-aligned model that separates bias-aligned  and bias-conflicting slices within each class. Conveniently, by simply ordering samples in ascending order of their likelihood of belonging to the ground truth class under the model, we can now identify bias-conflicting samples. While useful, this does not however discover coherent slices of data or segregate distinct majority (or minority) groups within a class when more than one exists. To address this, we perform Correlation-aware Slicing in the feature space of the bias-amplified model $h_{\sone}$.

\subsection{Then: (\underline{Co}rrelation-aware) \underline{S}l\underline{i}cing (\stwo)}
\label{sec:bamm}

To discover distinct and coherent bias-conflicting slices for each class, we fit a correlation-aware mixture in the bias-amplified feature space learned by $h_{\sone}$. To prevent any inter-class contamination of slices, we opt to fit a separate mixture model for each class.

\begin{figure*}[h]
\setlength{\tabcolsep}{0.62em}
\centering
\begin{floatrow}
\footnotesize
\capbtabbox[\FBwidth][][!htbp]   
{
\begin{tabular}{l c c c c c}
\toprule
 & \#Train  &  & \#BC  & Class &  \\ 
Dataset &  samples & \#Classes &  slices &  imbal. & Corr. (\%) \\
\midrule
Waterbirds &  4795& 2 & 2 & 2.9 & 95 \\
CelebA& 162770 & 2 & 1 & 5.7 & 98  \\
\nlo{}&  9349 & 6 & 30 & 2.9 & 75   \\
\nmi{}&  8209 & 6 & 30  & 3.5 & 90 \\
\nhi{}& 7839 & 6 & 30 & 3.8 & 95  \\
\bottomrule
\end{tabular}
}
{
\caption{\textbf{Summary of evaluation settings.} We study a diverse set of settings, including practical settings not considered in prior work: having \emph{no} \textit{bias-conflicting} slice for a class (CelebA), or having \emph{multiple} \textit{bias-conflicting} slices per-class (NICO++). Class imbalance is the ratio of the number of samples for the largest class to the smallest class. Correlation is the average \% of samples for a given spurious attribute that contain the correlated majority label. (BC=bias-conflicting, Class imbal.=Class imbalance, Corr.=Correlation).}
\label{tab:dataset}
}
\ffigbox[0.95\FBwidth][][!htbp]
{
\includegraphics[width=0.9\linewidth]{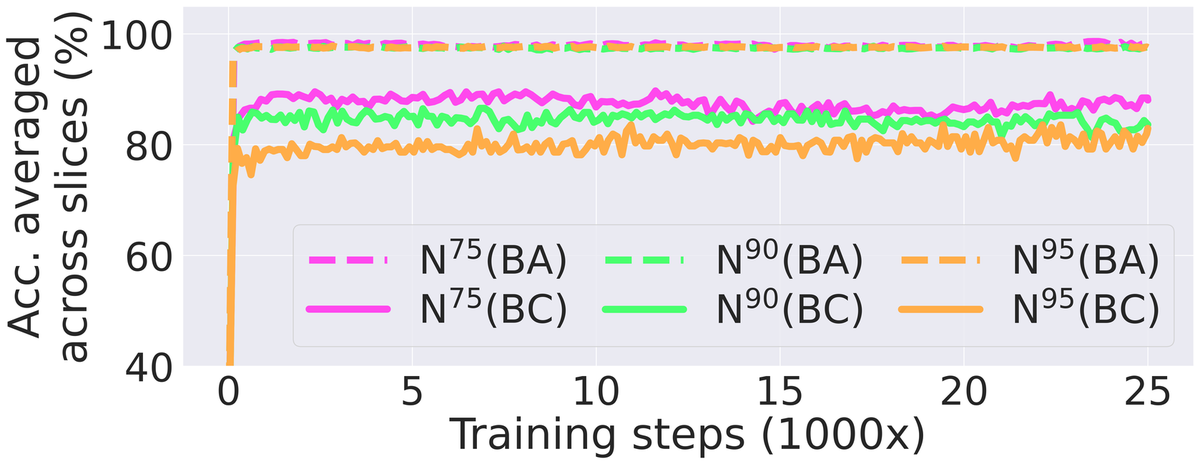}
}
{
\caption{\textbf{Effect of correlation strength.} We plot validation set accuracy of an ERM model trained on each of our proposed \nhi{}, \nmi{} and \nlo{} settings. Accuracies of \textit{bias-conflicting} (dashed) and \textit{bias-aligned} (solid) slices are shown separately. We observe that the accuracy gap between \textit{bias-aligned} and \textit{bias-conflicting} slices widens as the 
the correlation strength increases (BC=\textit{bias conflicting slices}, BA=\textit{bias aligned slices}, $\text{N}^x$=NICO$++^x$).}
\label{fig:nico_plus_plus}
}
\end{floatrow}
\end{figure*}

The mixture model assumes that the slice membership of samples is modeled using a categorical distribution $\hat{\textbf{S}} \thicksim Cat(\textbf{p})$ with parameters $\textbf{p}=[p_1, p_2 \dots p_{\hat{k}}]$, where $\sum_{i=1}^{\hat{k}} p_i = 1$. Here, $p_i$ gives the membership probability the mixture model assigns to the $i^{th}$ slice. We additionally model two sets of priors: 

\noindent i) \predterm, which we set to the predictive distribution (logits) from the biased model $h_\sone(x_i)$. The correlation prior encourages grouping together samples for which the biased model makes similar predictions; intuitively, for each class this corresponds to grouping together all bias-aligned (or all bias-conflicting samples). For each slice $S^{(j)}$, we model the biased predictions $B$ as a multivariate Gaussian distribution $\mathcal{N}(\mu_p^{(j)}, \Sigma_p^{(j)})$. 

\noindent (ii) \cohterm, which for sample $x_i$ is obtained using cross-modal (CLIP~\cite{radford2021learning}) embeddings $z_i = g(x_i)$, which enforces that slices correspond to semantically coherent concepts. Additionally, this prior also makes the slices discovered by our method amenable to automated captioning~\cite{mokady2021clipcap} (see Figure~\ref{fig:qualitative_results}) and prompt matching~\cite{eyuboglu2022domino,jain2022distilling} using CLIP. For each slice $S^{(j)}$ where $j = \{1,...,\hat{k}\}$, we fit a multivariate Gaussian distribution, $\mathcal{N}(\mu_c, \Sigma_c)$. 

Let $\textbf{m} = \left[\left\{\mu_p^{(j)}, \Sigma_p^{(j)}, \mu_c^{(j)}, \Sigma_c^{(j)}\right\}_{j=1}^{\hat{k}}\right]$ denote the set of all parameters of the multivariate distributions, where $\hat{k}$ is the number of slices predicted for each class. In total, we represent the full set of mixture model parameters for a class by $\phi=\left[\textbf{p}, \textbf{m} \right]$.
We fit $\phi$  so as to maximize the log-likelihood $l(\phi)$ on the validation set $D_{val}$ using expectation maximization (EM) ~\cite{dempster1977maximum}, which is given by:  
\begin{multline}\label{eq:BAMM_eq}
l(\phi) = \sum_{i = 1}^{|D_{val}|} \log \sum_{j = 1}^{\hat{k}} \left[ P(S^{(j)} = 1) P(Z = z_i|S^{(j)}=1) \right.\\
\left.P(B = h_b(x_i) | S^{(j)} = 1)^\alpha\right],
\end{multline}
\noindent and $\alpha$ controls the trade-off between the coherence and correlation priors. After fitting the per-class mixture models, we perform inference to
generate slice assignments, rank slices in increasing order of model accuracy, and present the top-k samples from the lowest performing slices to a practitioner for further intervention.

\section{Experiments}

In this section, we first describe our evaluation settings (Sec.~\ref{subsec:settings}). We then present details of baselines (Sec.~\ref{sec:baselines}) and quantitative results with our method (Sec.~\ref{subsec:results}), followed by an ablation study (Sec.~\ref{subsec:ablations}) and analysis (Sec.~\ref{subsec:analysis}). 

\subsection{Evaluation settings}
\label{subsec:settings}
We evaluate our method across diverse correlation  settings that vary in the number of classes, the number of bias-conflicting slices per class, the  degree of label imbalance across classes, and the strength of spurious correlation (measured by degree of group imbalance \emph{within} classes). We describe each setting below, and present summary statistics in Table~\ref{tab:dataset}.

\noindent\textbf{Waterbirds}~\cite{sagawa2020distributionally} consists of crops of landbirds and waterbirds~\cite{wah2011cub} superimposed on water and land backgrounds from the Places~\cite{zhou2017places} dataset. The task of interest is to distinguish between \emph{waterbirds} and \emph{landbirds}, where the background (\textit{water}/\textit{land}) is spuriously correlated with the label (\textit{waterbirds}/\textit{landbirds} respectively). The correlation biases here are $\textit{water} {\md}\textit{waterbirds}$ and $\textit{land}{\md}\textit{landbirds}$, resulting in two bias-conflicting slices (\textit{waterbirds in land} and \textit{landbirds in water}).

\noindent \textbf{CelebA}~\cite{liu2015faceattributes} contains images of celebrity faces annotated with various attributes such as gender, baldness, facial expression, and hair color. We use the entire training set of CelebA and consider the task of classifying images of people as either \textit{blonde} or \textit{not blonde}, using splits proposed in prior work on bias mitigation~\cite{liu2021jtt, li2023bias}. The training set has a \textit{male}${\md}$\textit{not blonde} spurious correlation resulting in a \emph{single} bias conflicting slice corresponding to \textit{blonde males}.

\noindent\textbf{NICO++}~\cite{zhang2022nico} consists of real-world images of concepts (eg. \emph{dog}, \emph{bike}, and \emph{wheat}) appearing in different contexts (eg. \emph{grass}, \emph{water}, and \emph{beach}). Context annotations for six contexts (\textit{dim lighting}, \textit{outdoor}, \textit{grass}, \textit{rock}, \textit{autumn}, \textit{water}) have been made public, and we use these to generate training, validation, and testing splits to simulate \emph{controlled} correlation settings for 6-way classification. We report on three such settings that we denote by \nhi{}, \nmi{} and \nlo{}, where the superscript denotes the degree to which each context is spuriously correlated with its corresponding class. As this correlation increases, the accuracy gap between bias-conflicting and bias-aligned slices of data widens (see Fig.~\ref{fig:nico_plus_plus}).

\begin{table*}[t]
\begin{center}
\begin{tabular}{lcccccc}
\toprule
Method  & Waterbirds & CelebA & \nlo{} & \nmi{} & \nhi{} \\
\midrule
FD~\cite{jain2022distilling}  &0.9& 0.7 & 0.19 & 0.19 & 0.19 \\
Domino~\cite{eyuboglu2022domino} & 1.0 & 0.9 & 0.24 & 0.25 &  0.27\\
\rowcolor{Gray}
\method (ours) & \textbf{1.0} & \textbf{0.9} & \textbf{0.56} & \textbf{0.60} & \textbf{0.62}\\
\bottomrule
\end{tabular}{
\vspace{-10pt}
\caption{\textbf{Our method results in better discovery of correlation slices across settings.} We report \paten (best=1.0) for retrieving ground truth bias-conflicting slices on the test set.}
\label{tab:prec_at_10_test}
}
\end{center}

\end{table*}

\noindent\textbf{Metrics.} Following Domino~\cite{eyuboglu2022domino}, we employ \pak for evaluation, which measures how accurately a slice discovery method is able to segregate the bias-conflicting slices. 
Assume $S = \{s_1, s_2, \dots s_l\}$ to be the set of ground truth bias-conflicting slices in a dataset $\mathcal{D}$. Let the slices predicted by an algorithm $A$ be $\hat{S} = \{\hat{s}_1, \hat{s}_2, \dots \hat{s}_m\}$. For a predicted slice $\hat{s}_j$, let $O_j$ = $\{o_{j1}, o_{j2} \dots o_{jn}\}$ give the sequence of sample indices ordered by decreasing likelihood of the sample $x$ belonging to the current slice. 
Given a ground truth slice $s_i$ and a predicted slice $\hat{s}_j$, we compute their similarity: $P_k(s_i, \hat{s}_j) = \frac{1}{k}\sum_{i=1}^{k} \mathbbm{1}[x_{o_{ji}} \in s_i]$. Each ground truth slice $s_i$ is then mapped to the most similar predicted slice using $\argmax_{s\in \hat{S}_j} P_k(s_i, s)$. We then average the similarity score between the ground truth slices and their best-matching predicted slices. Specifically, for a slice discovery algorithm $A$ we compute: 
$$\pak(A) = \frac{1}{l}\sum_{i=1}^{l} \underset{j \in [m]}{\max} P_k(s_i, \hat{s}_j) $$

In addition, to solely evaluate the effectiveness of the first phase of our method (Sec.~\ref{sec:bai}) at ranking samples in order of their bias alignment, we compute \avgap: For each class containing a bias-conflicting slice, we compute the Average Precision (AP) score that measures how good a given ranking is at separating bias-conflicting samples from the bias-aligned samples. We then average across classes to obtain \avgap.

\noindent\textbf{Implementation details.}
For all datasets, we train ResNet50~\cite{he2016deep} models  using an SGD optimizer with a momentum of 0.9 and batch size of 64 starting from supervised ImageNet~\cite{krizhevsky2017imagenet} initialisations. Following~\cite{liu2021jtt}, we train models for 300 epochs on Waterbirds and 50 epochs on CelebA. On NICO++, we train for 25k steps. For CelebA and Waterbirds, we follow prior work~\cite{liu2021jtt} and use a learning rate of $10^{-3}$ and weight decay of $10^{-4}$ for training the ERM models used by our baselines. We use a learning rate of $10^{-5}$ for training bias-amplified models $h_\sone$. For weight decay, we sweep over the range $[10^{-3}, 10^{-2}, 10^{-1}, 1.0, 2.0]$ using the strategy described in Sec.~\ref{sec:bai}.
For \stwo, we run expectation maximization~\cite{dempster1977maximum} for 100 steps or until the log-likelihood increases by less than $10^{-7}$ in successive steps. We set the number of slices per class, $\hat{k}$ to a large value of $36$. We use a full covariance matrix for modeling the covariance of $B \mid S^{j}{=}1$ ($\Sigma_p$), while we follow prior work~\cite{eyuboglu2022domino} to restrict the covariance matrix of $Z \mid S^{j}{=}1$ ($\Sigma_c$) to be diagonal. Further, we use a non-negative regularization of $\delta_p$ on the diagonal elements of $\Sigma_p$. %

\subsection{Baselines} 
\label{sec:baselines}

\noindent We compare against the following slice discovery methods:

\noindent i) \textbf{Domino}~\cite{eyuboglu2022domino} discovers slices by fitting an ``error-aware'' mixture model to a combination of multi-modal (CLIP~\cite{radford2021learning}) embeddings of validation set examples, ERM model predictions, and labels. While the second phase of our method is motivated by Domino, there are some important distinctions: Domino does \emph{not} use bias amplification, fits a single mixture model \emph{across classes} with an additional \emph{soft} constraint on class membership, and models a categorical distribution over the predicted labels 
(rather than using the full set of logits from the bias-amplified model, see Sec.~\ref{sec:bamm}).

\noindent ii) \textbf{Failure Directions (FD)}~\cite{jain2022distilling} trains an SVM (per-class) to predict whether a standard ERM model would misclassify a given validation sample. The distance to the per-class SVM boundary is then used to score samples on an evaluation set. However, unlike \method, this method is constrained to retrieving only a \emph{single} failure mode per-class.

\subsection{Results}
\label{subsec:results}

In Table~\ref{tab:prec_at_10_test}, we compare the performance of our method to Domino~\cite{eyuboglu2022domino} and Failure Directions~\cite{jain2022distilling} (FD) for discovering bias-conflicting slices on the test set of the Waterbirds, CelebA, and NICO++ datasets. Following Domino, we measure precision at $k=10$. A high-precision slice of 10 images will likely be representative of a given bias-conflicting mode while still being of manageable size to be presented to a practitioner for intervention. 

On Waterbirds and CelebA, our method achieves \pak of 1.0 and 0.9 respectively on the test set, matching prior work \cite{eyuboglu2022domino}. 

On the other hand, we see significant gains (\textbf{$>$+0.37}) on the more challenging NICO++ settings. Recall that these settings contain a controlled degree of correlation with multiple (5) bias-conflicting slices per class. As expected, FD~\cite{jain2022distilling} underperforms in this challenging setting as it cannot generalize to $>1$ minority group per-class. Somewhat surprisingly, we find that Domino~\cite{eyuboglu2022domino} also underperforms in this setting despite containing a clustering phase, presumably  because it does not operate in bias-amplified feature space, models a less informative prior, and employs a soft class assignment.

\begin{figure}[h]
    \centering
\includegraphics[width=\textwidth]{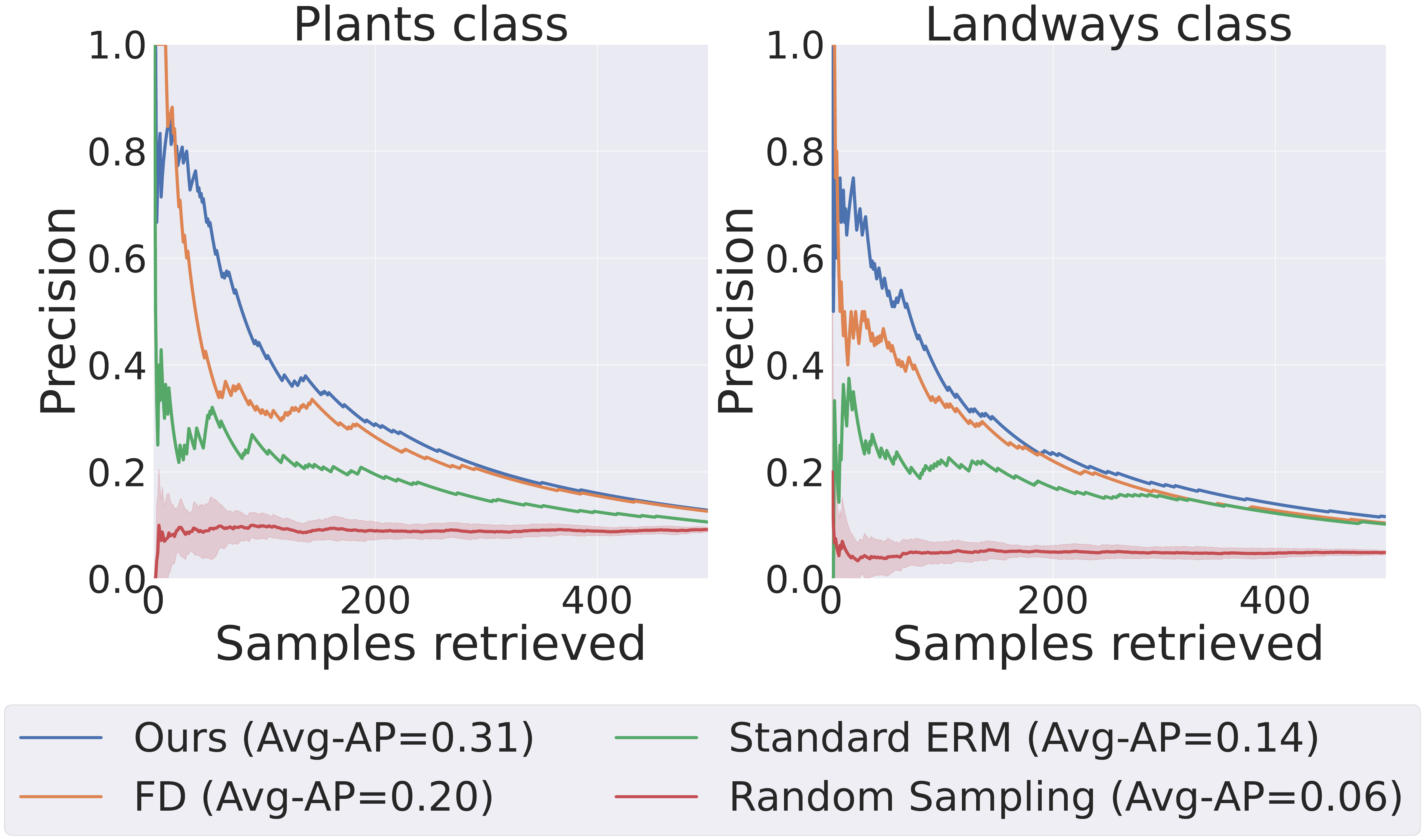}
    \caption{\textbf{Classwise Precision for retrieving bias-conflicting samples} from Plants and Landways classes in \nhi{} setting as the number of retrieved samples changes. The legend gives the \avgap scores for the different methods being compared.}
    \label{fig:prec-at-k}
\end{figure}
\noindent\textbf{Evaluating Bias Identification.}
We now evaluate the effectiveness of the first stage of our method, \sone at simply \emph{identifying} bias-conflicting samples, without having to separate them into distinct categories. Recall that \sone can conveniently identify bias-conflicting samples by retrieving samples that have a low likelihood of belonging to the ground truth class under the bias-amplified model. To do so, we follow~\cite{jain2022distilling} and plot per-class precision at retrieving bias-conflicting samples for two classes in \nhi{}, as a function of the number of samples retrieved. We compare against Failure Directions~\cite{jain2022distilling} (FD), using per-sample ground truth confidences under a \emph{standard} ERM model, and random sampling from the base population. 

Figure~\ref{fig:prec-at-k} shows results.  For both \emph{plants} and \emph{landways} classes, we see that for retrieving more than a few samples, \sone consistently outperforms all methods, and obtains the best \avgap (\textbf{+0.11} higher than the next best).

\subsection{Ablations}
\label{subsec:ablations}

\noindent\textbf{Ablating \sone.}
We now ablate the first stage of our proposed method and report \avgap on \nhi. We find:

\noindent\textbf{$\triangleright$ Amplifying Correlations improves retrieval (Table~\ref{tab:ablate_com_att}, Row 1 v/s 5).} We first compare our method against retrieving samples with a low confidence of belonging to the ground truth class under a standard ERM model, without any amplification. We find that bias amplification \emph{significantly} boosts retrieval (\textbf{+0.17} \avgap), validating our hypothesis that such amplification increases feature separation.

\noindent\textbf{$\triangleright$ Our proposed amplification strategy outperforms competing  methods (Table~\ref{tab:ablate_com_att}, Row 2-4 v/s 5).} We compare against alternative loss objectives: i) GCE~\cite{nam2020learning}, ii) Li~\emph{et al.} that combine a squared loss with an additional auxiliary variable, iii) A simple linear probing~\cite{ye2019unsupervised} amplification strategy where we update only the classifier head and freeze all other parameters. On \nhi, we observe that our strategy outperforms the next best BA method by 
\textbf{+0.06}. 

\begin{table}[h]    
  \centering
  \RawFloats    
  \begin{subfloatrow}
  \footnotesize
  \ffigbox[\FBwidth][][b]{
    \setlength{\tabcolsep}{2pt}      
      \begin{tabular}{l c}
        \toprule
        \textbf{Method} & \avgap \\
        \midrule
        None (ERM) & 0.14 \\
        GCE~\cite{nam2020learning}  & 0.13 \\       
        Linear probe  & 0.25 \\       
        Sq. loss + Aux var.~\cite{li2023bias}  & 0.22 \\             
        \rowcolor{Gray}
        Ours  & \textbf{0.31} \\  
        \bottomrule
        \end{tabular}
}
{
    \caption{\small Varying amplification.}
    \label{tab:ablate_com_att}
  }
\hspace{-1.0cm}
\ffigbox[\FBwidth][][]
{
      \setlength{\tabcolsep}{2pt}
      \begin{tabular}{lc}
        \toprule
        \textbf{Method} & \avgap \\
        \midrule        
        Max val ClassDiff~\cite{li2023bias} & 0.06 \\       
        Max val acc. (Ours) & \textbf{0.31} \\     
        \rowcolor{Gray}
        Max train acc. (Ours) & \textbf{0.31} \\          
        \midrule
        Oracle & 0.36 \\
        \bottomrule
        \end{tabular}
}
  {
  \caption{\small Varying stopping criterion.}\label{tab:ablate_stopping}
  }
  \end{subfloatrow}
    \caption{\textbf{Ablating \sone}. We report \avgap at retrieving bias-conflicting samples on the \nhi{} train set.}
  \vspace{20pt}
\end{table}

\begin{figure*}[ht]
    \centering
    \includegraphics[width=0.96\linewidth]{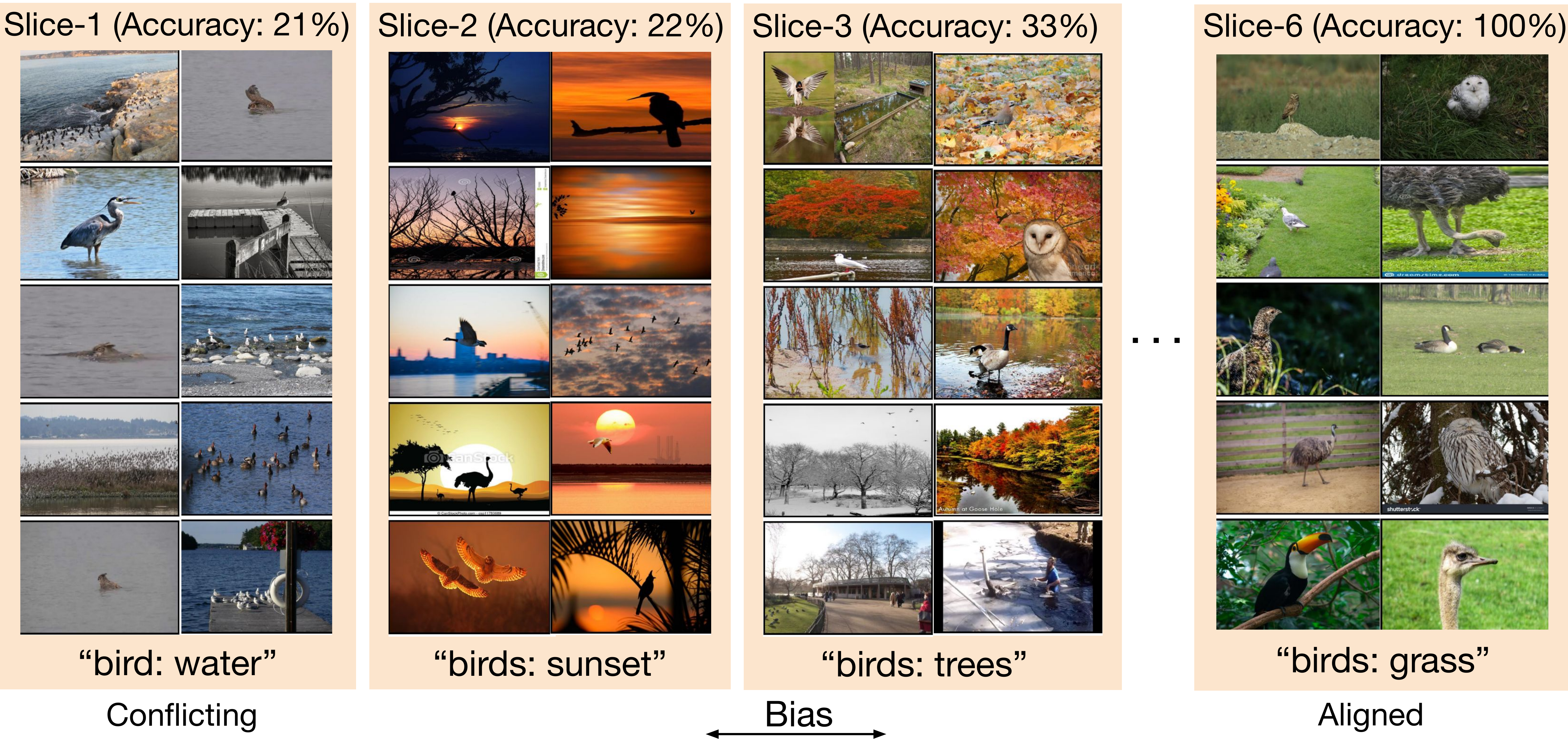}
    \vspace{-6pt}
    \caption{\textbf{Slices discovered by \method.} We present the top-10 samples in  slices predicted by our method for the \textit{bird} class from \nhi{}, ordered in increasing order of model accuracy. The ground truth correlation 
    in this case is $\textit{grass} {\md}\textit{bird}$, and so our goal is to discover dataset slices wherein this correlation \emph{does not hold}. As shown, the first two slices discovered by \method correspond to two distinct and coherent bias-conflicting slices -- \emph{bird in water} and \emph{bird at sunset} (slice name obtained by extracting frequent keywords from a generated caption  ~\cite{mokady2021clipcap}).
    On the right we show an example of a \emph{bias-aligned} slice discovered by our method, which may easily be discarded by filtering based on accuracy.}
    \label{fig:qualitative_results}
\end{figure*}

\begin{table}[h]
\begin{center}
\begin{tabular}{lc}
\toprule
Objective  & \paten \\
\midrule
CLIP & 0.38\\
Predicted logits   & 0.58\\
Predicted label & 0.26\\
CLIP + Predicted label~\cite{eyuboglu2022domino} & 0.38\\
\rowcolor{Gray}
CLIP + Predicted logits (Ours) & \textbf{0.62} \\
\bottomrule
\end{tabular}
\end{center}
{
\vspace{-10pt}
\caption{\textbf{Ablating \stwo.} We report \paten at identifying bias-conflicting slices on \nhi.}
\label{tab:bamm_objective}
}
\end{table}

\begin{figure}
    \centering
\includegraphics[width=\textwidth]{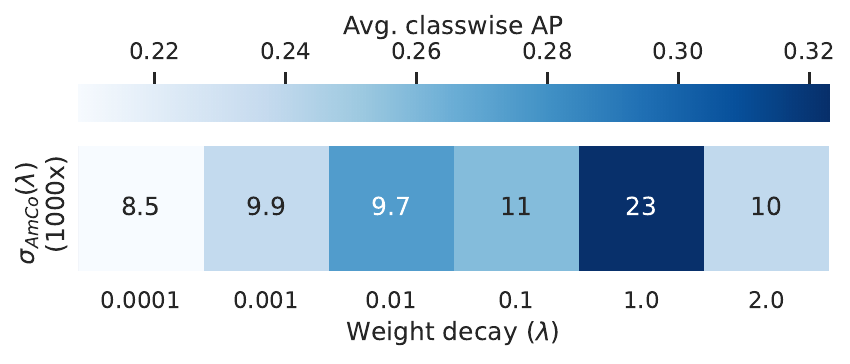}
\vspace{-15pt}
    \caption{\textbf{Validating $\lambda$ tuning  strategy.} We plot the \avgap (color, darker is higher) across $\lambda$ values along with the corresponding variance in confidence \varconf (inscribed values), at the point where training accuracy peaks. As seen, this best \avgap is achieved at the $\lambda$ that maximizes \varconf. 
    }
    \label{fig:wd_tune}
\end{figure}
\vspace{-20pt}
\noindent \textbf{$\triangleright$ Maximum training accuracy is an effective stopping criterion. (Table~\ref{tab:ablate_stopping}})
Recall that for retrieving bias-conflicting samples, we seek to identify a snapshot in training at which bias amplification is large, and in our experiments we pick the point at which training accuracy peaks. To evaluate whether this is a reasonable stopping criterion, we compare against alternatives: i) maximum validation accuracy, ii) maximum ClassDiff, adapted from Li~\emph{et al.}~\cite{li2023bias} which finds the average difference in model validation accuracy between all class pairs (ClassDiff), to be strongly anti-correlated with worst-group accuracy, following which we pick the point with the highest ClassDiff / lowest worst-group accuracy, indicating maximum amplification, and iii) an oracle strategy that picks a model corresponding to the best retrieval. We find that using maximum training or validation accuracy both perform well and reach within 0.05 AP of oracle retrieval. %

\noindent\textbf{Ablating \stwo.} Now
In Table~\ref{tab:bamm_objective} we try variations of our proposed mixture modeling strategy \stwo, and report \paten on \nhi. We find that dropping either term of our objective in Eq.~\ref{eq:BAMM_eq} (distribution on logits within a slice or distribution on embeddings within a slice) degrades precision. Further, we compare with using only the predicted label rather than the complete logits distribution as done in Domino~\cite{eyuboglu2022domino}; we find that this \emph{signficantly} drops performance (\textbf{-0.24} \paten). We hypothesize that this is because the logits distribution is more informative, which may be helpful in identifying bias-conflicting samples on which the model does not fail.

\subsection{Analysis}
\label{subsec:analysis}

\noindent \textbf{Qualitative results.} While our method uncovers coherent slices corresponding to dataset correlations, in practice it is important to be able to identify a small subset of \emph{coherent, bias-conflicting} slices that may be presented to a practitioner. To do so, we order slices by validation accuracy (least to most). 
Fig.~\ref{fig:qualitative_results} shows slices discovered by our method on the \textit{bird} class of \nhi, which is spuriously correlated with the \textit{grass} context. For each predicted slice, we  visualize the top-10 samples ranked based on their likelihood of belonging to the slice. We observe that \method is able to uncover coherent bias-conflicting slices, which we name using the strategy described in Kim \emph{et al.}~\cite{kim2023explaining} to obtain slice descriptions like \textit{birds: water}, \textit{birds: sunset} and so on. We note that these bias-conflicting slices also correspond to failure modes for the model (evidenced by low accuracy). On the right, we visualize a bias-aligned slice discovered by \method (\textit{birds in grass} (\#6)), which can be easily filtered out based on its high validation accuracy (in this case, 100\%).

\looseness=-1 \noindent \textbf{Measuring bias amplification.} To verify that our bias-amplification strategy indeed results in learning a bias-aligned decision boundary, we compute the difference in training accuracies of the actual bias-aligned  and bias-conflicting slices per class (\texttt{GT-Acc-gap}). On \nhi{}, we find that \texttt{GT-Acc-gap} strongly correlates with \avgap on \nhi{} (Pearson coefficient $0.85$), verifying that bias amplification does indeed lead to better retrieval. 
Further, our bias amplified model $h_{\sone}$ achieves a \texttt{GT-Acc-gap} of 0.23 (at the maximum training accuracy), while a standard ERM model achieves \texttt{GT-Acc-gap} of 0.001 (at its maximum training accuracy): clearly, our \sone strategy successfully amplifies correlations. 

\noindent\textbf{Validating weight decay tuning strategy.}
Recall from Section ~\ref{sec:bai} that we choose the weight decay value that maximizes the per-class average of variance in the likelihood of the ground truth label under the model, at the point where the training accuracy is maximum (\varconf). In Fig.~\ref{fig:wd_tune}, we compute \varconf across a range of $\lambda$ values and verify that the best \avgap is indeed achieved at the maximum \varconf.

\begin{figure}[t!]
    \centering
    \begin{subfigure}[b]{0.5\textwidth}
        \centering
        \includegraphics[width=\textwidth]{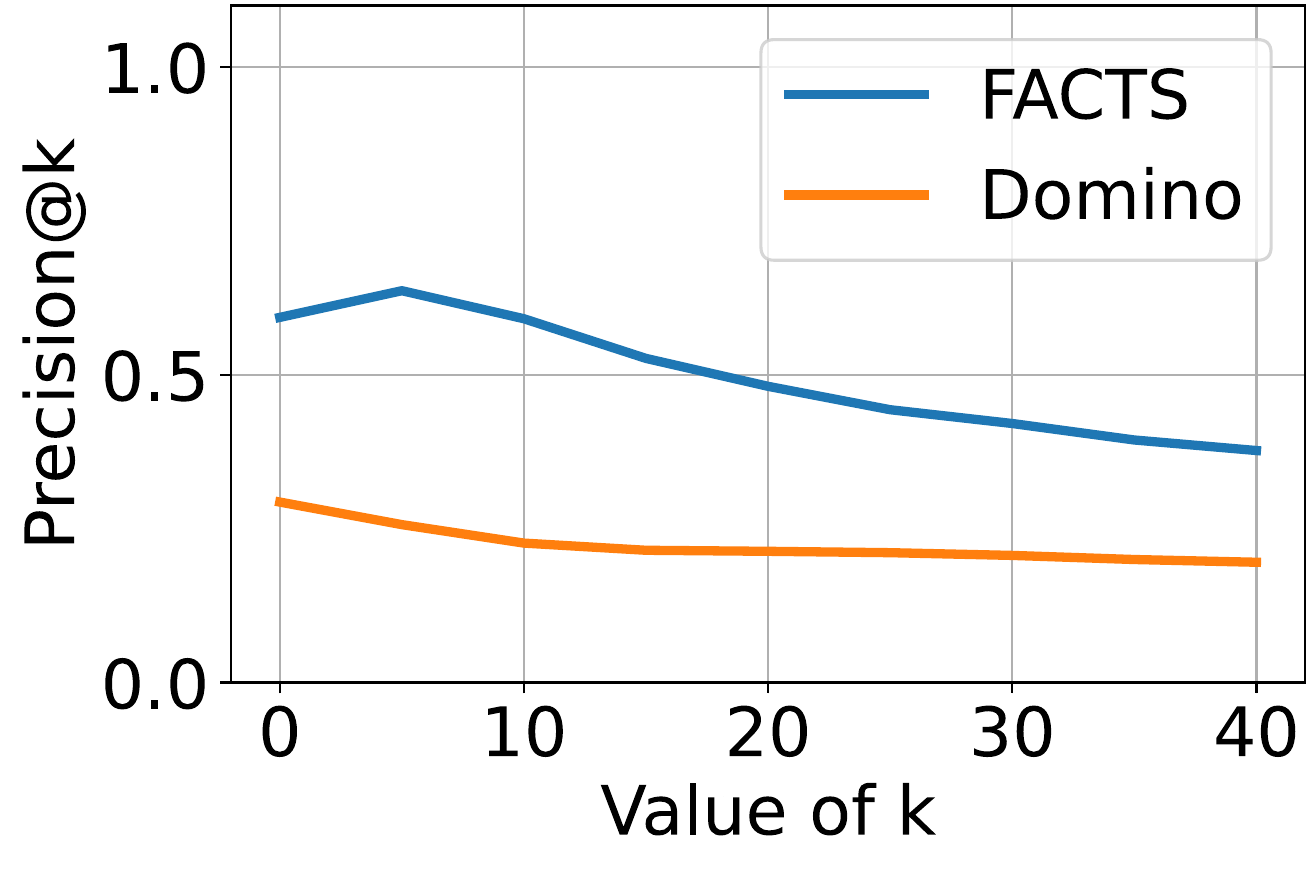}
        \caption{\nhi{}}
    \end{subfigure}%
    ~ 
    \begin{subfigure}[b]{0.5\textwidth}
        \centering
        \includegraphics[width=\textwidth]{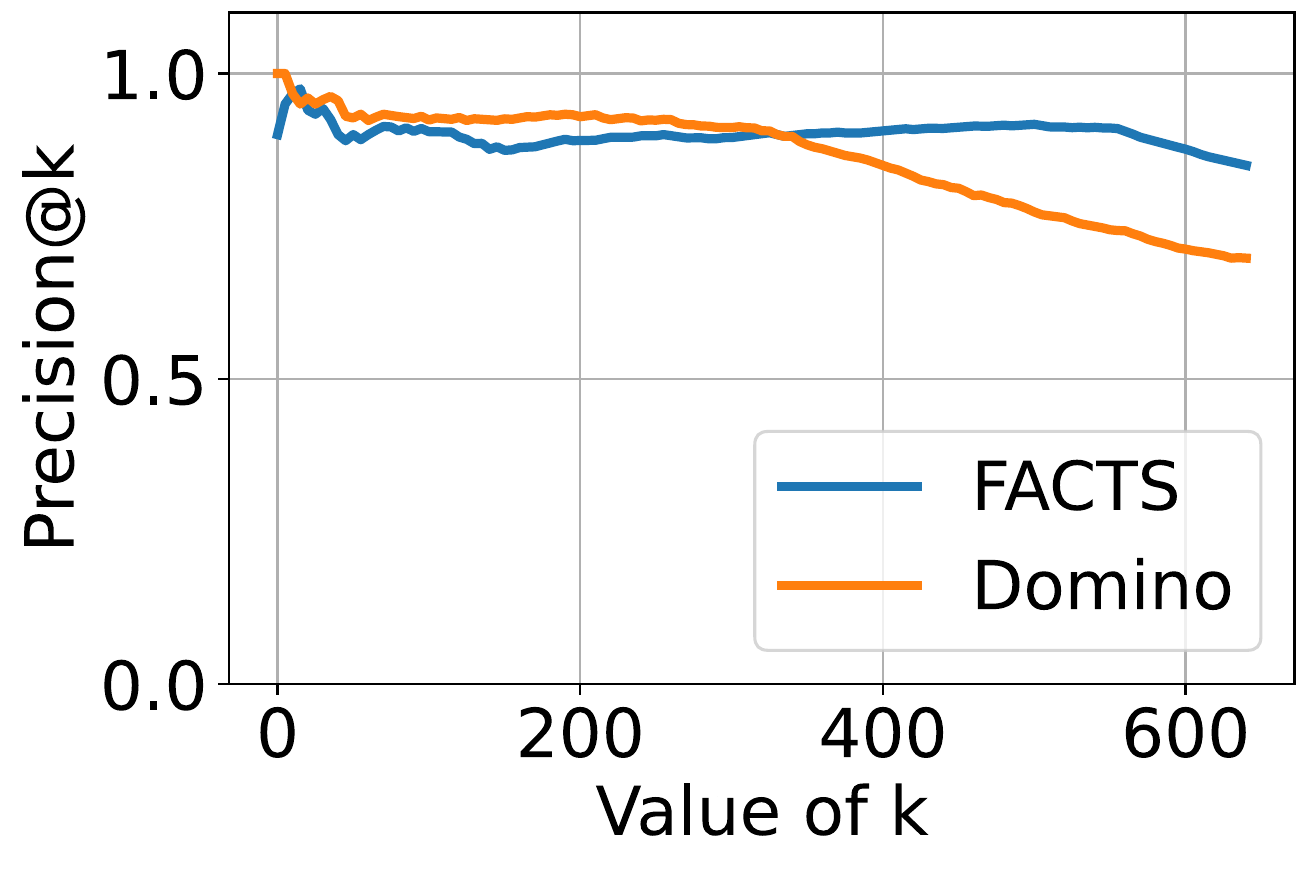}
        \caption{Waterbirds}
    \end{subfigure}
    \caption{
    \textbf{\pak for different values of $k$.} We plot the \pak obtained by Domino~\cite{eyuboglu2022domino} and \method. as $k$ varies for \nhi{} and Waterbirds. }
    \label{fig:pak}
\end{figure}

\noindent \textbf{\pak for different values of $k$.} We perform our primary evaluations (Table~\ref{tab:prec_at_10_test}) using the \paten metric (Sec.~\ref{subsec:settings}). We additionally benchmark \pak across a range of values of $k$ in Fig.~\ref{fig:pak}. For Waterbirds, \pak of \method remains relatively stable, even for large values of $k$ that match the total number of bias-conflicting samples per slice ($642$). In the case of \nhi{}, we observe that \method performs better than Domino~\cite{eyuboglu2022domino} across different values of $k$.

\noindent\textbf{Additional slice evaluation metrics.} 
We evaluate the quality of slices predicted by different methods using a few additional metrics in Table~\ref{tab:additional_metrics}. For each ground truth slice with $\beta$ samples, we compute Recall@$\beta$ and AP score obtained by the best matching predicted slice (see~\ref{subsec:settings} for details on how ground truth slices are associated to predicted slices). We average these metrics across slices to obtain Avg. Slice Recall and Avg. Slice AP. As seen, \method predicts slices that achieve better recall and AP compared to Domino~\cite{eyuboglu2022domino}.

\noindent\textbf{Evaluating slice ordering.}
The $\pak$ metric (proposed by Domino~\cite{eyuboglu2022domino}) only evaluates the ability of slicing algorithms to predict a matching slice for every ground truth slice. However, it does not evaluate their ability to separate the bias-conflicting slices from bias-aligned slices. As in Figure~\ref{fig:approach_fig}, we order the predicted slices by increasing validation accuracy. We additionally call a predicted slice as "bias-conflicting" if at least 6 out of the top-10 samples of that slice are bias-conflicting. We use Slice Ranking AP to evaluate the extent to which the ordering places slices marked as "bias-conflicting" on top. We observe that bias-conflicting slices predicted by \method can be  better separated using validation accuracy when compared to Domino~\cite{eyuboglu2022domino} (Table~\ref{tab:additional_metrics}).
   
\begin{table}[h!]
    \centering
    \resizebox{0.95\columnwidth}{!}{
    \begin{tabular}{l c c c}
    \toprule
         Method & Avg. Slice  & Avg. Slice  & Slice  \\
         & Recall &AP & Ranking AP\\
    \midrule
    Domino~\cite{eyuboglu2022domino} & 0.15 & 0.24 & 0.92\\
    \rowcolor{Gray}\texttt{FACTS} & \textbf{0.34} & \textbf{0.40} & \textbf{0.97}\\

    \bottomrule
    \end{tabular}
    }
    \caption{\textbf{Additional metrics.} In addition to \pak (\cite{eyuboglu2022domino}), we use Avg. Slice Recall/AP and Slice Ranking AP to evaluate the slice quality and slice ordering respectively on \nhi{}.}
    \label{tab:additional_metrics}
\end{table}

In the supplementary, we present additional analysis for the sensitivity of our clustering strategy to its hyperparameters (we find it to be relatively stable), as well as additional qualitative results and ablation studies.

\section{Discussion}

We propose novel algorithm for automatically discovering correlation bias, that first amplifies correlations to learn a context-aligned decision boundary and then clusters instances in this bias-amplified feature space to uncover semantically coherent bias-conflicting slices. We show that despite its simplicity, our method significantly outperforms prior work across a diverse set of evaluation settings.

\noindent\textbf{Limitations.} We focus on identifying only correlation bias, and leave the discovery of other types of bias (\eg due to mislabeled or ambiguous examples) to future work. Further, we make certain simplifying assumptions, \eg a datapoint does not contain multiple spurious attributes. Finally, we restrict experiments to ResNet50~\cite{he2016deep} architectures trained with SGD in this work. Despite these limitations, we envisage that our work will be useful in identifying spurious correlations across a broad range of real-world settings.

\section{Acknowledgements}
 This work was supported in part by funding from Cisco, Google, DARPA LwLL, NASA ULI, and NSF \#2144194. We thank George Stoica, Simar Kareer, Aaditya Singh and Sahil Khose for their feedback on the first draft of this work. We additionally thank Simar Kareer for help with figures.

{\small
\bibliographystyle{unsrt}
\bibliography{main}

\begin{thebibliography}{10}

\bibitem{torralba2011unbiased}
Antonio Torralba and Alexei~A Efros.
\newblock Unbiased look at dataset bias.
\newblock In {\em CVPR 2011}, pages 1521--1528. IEEE, 2011.

\bibitem{caliskan2017semantics}
Aylin Caliskan, Joanna~J Bryson, and Arvind Narayanan.
\newblock Semantics derived automatically from language corpora contain
  human-like biases.
\newblock {\em Science}, 356(6334):183--186, 2017.

\bibitem{buolamwini2018gender}
Joy Buolamwini and Timnit Gebru.
\newblock Gender shades: Intersectional accuracy disparities in commercial
  gender classification.
\newblock In {\em Conference on fairness, accountability and transparency},
  pages 77--91. PMLR, 2018.

\bibitem{vapnik1999nature}
Vladimir Vapnik.
\newblock {\em The nature of statistical learning theory}.
\newblock Springer science \& business media, 1999.

\bibitem{geirhos2020shortcut}
Robert Geirhos, J{\"o}rn-Henrik Jacobsen, Claudio Michaelis, Richard Zemel,
  Wieland Brendel, Matthias Bethge, and Felix~A Wichmann.
\newblock Shortcut learning in deep neural networks.
\newblock {\em Nature Machine Intelligence}, 2(11):665--673, 2020.

\bibitem{jabbour2020-pmlr-v126}
Sarah Jabbour, David Fouhey, Ella Kazerooni, Michael~W. Sjoding, and Jenna
  Wiens.
\newblock Deep learning applied to chest x-rays: Exploiting and preventing
  shortcuts.
\newblock In Finale Doshi-Velez, Jim Fackler, Ken Jung, David Kale, Rajesh
  Ranganath, Byron Wallace, and Jenna Wiens, editors, {\em Proceedings of the
  5th Machine Learning for Healthcare Conference}, volume 126 of {\em
  Proceedings of Machine Learning Research}, pages 750--782. PMLR, 07--08 Aug
  2020.

\bibitem{badgeley2019deep}
Marcus~A Badgeley, John~R Zech, Luke Oakden-Rayner, Benjamin~S Glicksberg,
  Manway Liu, William Gale, Michael~V McConnell, Bethany Percha, Thomas~M
  Snyder, and Joel~T Dudley.
\newblock Deep learning predicts hip fracture using confounding patient and
  healthcare variables.
\newblock {\em NPJ digital medicine}, 2(1):31, 2019.

\bibitem{winkler2019association}
Julia~K Winkler, Christine Fink, Ferdinand Toberer, Alexander Enk, Teresa
  Deinlein, Rainer Hofmann-Wellenhof, Luc Thomas, Aimilios Lallas, Andreas
  Blum, Wilhelm Stolz, et~al.
\newblock Association between surgical skin markings in dermoscopic images and
  diagnostic performance of a deep learning convolutional neural network for
  melanoma recognition.
\newblock {\em JAMA dermatology}, 155(10):1135--1141, 2019.

\bibitem{wilson2019predictive}
Benjamin Wilson, Judy Hoffman, and Jamie Morgenstern.
\newblock Predictive inequity in object detection.
\newblock {\em arXiv preprint arXiv:1902.11097}, 2019.

\bibitem{nam2022spread}
Junhyun Nam, Jaehyung Kim, Jaeho Lee, and Jinwoo Shin.
\newblock Spread spurious attribute: Improving worst-group accuracy with
  spurious attribute estimation.
\newblock In {\em International Conference on Learning Representations}, 2022.

\bibitem{sagawa2020distributionally}
Shiori Sagawa*, Pang~Wei Koh*, Tatsunori~B. Hashimoto, and Percy Liang.
\newblock Distributionally robust neural networks.
\newblock In {\em International Conference on Learning Representations}, 2020.

\bibitem{liu2021jtt}
Evan~Z Liu, Behzad Haghgoo, Annie~S Chen, Aditi Raghunathan, Pang~Wei Koh,
  Shiori Sagawa, Percy Liang, and Chelsea Finn.
\newblock Just train twice: Improving group robustness without training group
  information.
\newblock In Marina Meila and Tong Zhang, editors, {\em Proceedings of the 38th
  International Conference on Machine Learning}, volume 139 of {\em Proceedings
  of Machine Learning Research}, pages 6781--6792. PMLR, 18--24 Jul 2021.

\bibitem{nam2020learning}
Junhyun Nam, Hyuntak Cha, Sungsoo Ahn, Jaeho Lee, and Jinwoo Shin.
\newblock Learning from failure: De-biasing classifier from biased classifier.
\newblock {\em Advances in Neural Information Processing Systems},
  33:20673--20684, 2020.

\bibitem{li2023bias}
Gaotang Li, Jiarui Liu, and Wei Hu.
\newblock Bias amplification improves worst-group accuracy without group
  information, 2023.

\bibitem{jain2022distilling}
Saachi Jain, Hannah Lawrence, Ankur Moitra, and Aleksander Madry.
\newblock Distilling model failures as directions in latent space.
\newblock In {\em ArXiv preprint arXiv:2206.14754}, 2022.

\bibitem{chawla2002smote}
Nitesh~V Chawla, Kevin~W Bowyer, Lawrence~O Hall, and W~Philip Kegelmeyer.
\newblock Smote: synthetic minority over-sampling technique.
\newblock {\em Journal of artificial intelligence research}, 16:321--357, 2002.

\bibitem{he2008adasyn}
Haibo He, Yang Bai, Edwardo~A Garcia, and Shutao Li.
\newblock Adasyn: Adaptive synthetic sampling approach for imbalanced learning.
\newblock In {\em 2008 IEEE international joint conference on neural networks
  (IEEE world congress on computational intelligence)}, pages 1322--1328. IEEE,
  2008.

\bibitem{mokady2021clipcap}
Ron Mokady, Amir Hertz, and Amit~H Bermano.
\newblock Clipcap: Clip prefix for image captioning.
\newblock {\em arXiv preprint arXiv:2111.09734}, 2021.

\bibitem{wong2021leveraging}
Eric Wong, Shibani Santurkar, and Aleksander Madry.
\newblock Leveraging sparse linear layers for debuggable deep networks.
\newblock In Marina Meila and Tong Zhang, editors, {\em Proceedings of the 38th
  International Conference on Machine Learning}, volume 139 of {\em Proceedings
  of Machine Learning Research}, pages 11205--11216. PMLR, 18--24 Jul 2021.

\bibitem{singla2022salient}
Sahil Singla and Soheil Feizi.
\newblock Salient imagenet: How to discover spurious features in deep learning?
\newblock In {\em International Conference on Learning Representations}, 2022.

\bibitem{shah2022modeldiff}
Harshay Shah, Sung~Min Park, Andrew Ilyas, and Aleksander Madry.
\newblock Modeldiff: A framework for comparing learning algorithms.
\newblock {\em arXiv preprint arXiv:2211.12491}, 2022.

\bibitem{oakden2020meaningfulfailures}
Luke Oakden-Rayner, Jared Dunnmon, Gustavo Carneiro, and Christopher R{\'e}.
\newblock Hidden stratification causes clinically meaningful failures in
  machine learning for medical imaging.
\newblock In {\em Proceedings of the ACM conference on health, inference, and
  learning}, pages 151--159, 2020.

\bibitem{krishnakumar2021udis}
Arvindkumar Krishnakumar, Viraj Prabhu, Sruthi Sudhakar, and Judy Hoffman.
\newblock Udis: Unsupervised discovery of bias in deep visual recognition
  models.
\newblock In {\em British Machine Vision Conference (BMVC)}, volume~1, page~3,
  2021.

\bibitem{eyuboglu2022domino}
Sabri Eyuboglu, Maya Varma, Khaled~Kamal Saab, Jean{-}Benoit Delbrouck,
  Christopher Lee{-}Messer, Jared Dunnmon, James Zou, and Christopher R{\'{e}}.
\newblock Domino: Discovering systematic errors with cross-modal embeddings.
\newblock In {\em The Tenth International Conference on Learning
  Representations, {ICLR} 2022, Virtual Event, April 25-29, 2022}.
  OpenReview.net, 2022.

\bibitem{zhang2023diagnosing}
Yuhui Zhang, Jeff~Z. HaoChen, Shih-Cheng Huang, Kuan-Chieh Wang, James Zou, and
  Serena Yeung.
\newblock Diagnosing and rectifying vision models using language.
\newblock In {\em The Eleventh International Conference on Learning
  Representations}, 2023.

\bibitem{kim2023explaining}
Younghyun Kim, Sangwoo Mo, Minkyu Kim, Kyungmin Lee, Jaeho Lee, and Jinwoo
  Shin.
\newblock Explaining visual biases as words by generating captions.
\newblock {\em arXiv preprint arXiv:2301.11104}, 2023.

\bibitem{bansal2022measures}
Rachit Bansal, Danish Pruthi, and Yonatan Belinkov.
\newblock Measures of information reflect memorization patterns.
\newblock In Alice~H. Oh, Alekh Agarwal, Danielle Belgrave, and Kyunghyun Cho,
  editors, {\em Advances in Neural Information Processing Systems}, 2022.

\bibitem{liu2015faceattributes}
Ziwei Liu, Ping Luo, Xiaogang Wang, and Xiaoou Tang.
\newblock Deep learning face attributes in the wild.
\newblock In {\em Proceedings of International Conference on Computer Vision
  (ICCV)}, December 2015.

\bibitem{zhang2022nico}
Xingxuan Zhang, Yue He, Renzhe Xu, Han Yu, Zheyan Shen, and Peng Cui.
\newblock Nico++: Towards better benchmarking for domain generalization, 2022.

\bibitem{singla2021understanding}
Sahil Singla, Besmira Nushi, Shital Shah, Ece Kamar, and Eric Horvitz.
\newblock Understanding failures of deep networks via robust feature
  extraction.
\newblock In {\em Proceedings of the IEEE/CVF Conference on Computer Vision and
  Pattern Recognition}, pages 12853--12862, 2021.

\bibitem{radford2021learning}
Alec Radford, Jong~Wook Kim, Chris Hallacy, Aditya Ramesh, Gabriel Goh,
  Sandhini Agarwal, Girish Sastry, Amanda Askell, Pamela Mishkin, Jack Clark,
  et~al.
\newblock Learning transferable visual models from natural language
  supervision.
\newblock In {\em International conference on machine learning}, pages
  8748--8763. PMLR, 2021.

\bibitem{paranjape2023agro}
Bhargavi Paranjape, Pradeep Dasigi, Vivek Srikumar, Luke Zettlemoyer, and
  Hannaneh Hajishirzi.
\newblock {AGRO}: Adversarial discovery of error-prone groups for robust
  optimization.
\newblock In {\em International Conference on Learning Representations}, 2023.

\bibitem{zhou22d-pmlr-v162}
Xiao Zhou, Yong Lin, Renjie Pi, Weizhong Zhang, Renzhe Xu, Peng Cui, and Tong
  Zhang.
\newblock Model agnostic sample reweighting for out-of-distribution learning.
\newblock In Kamalika Chaudhuri, Stefanie Jegelka, Le~Song, Csaba Szepesvari,
  Gang Niu, and Sivan Sabato, editors, {\em Proceedings of the 39th
  International Conference on Machine Learning}, volume 162 of {\em Proceedings
  of Machine Learning Research}, pages 27203--27221. PMLR, 17--23 Jul 2022.

\bibitem{hall2022bias}
Melissa Hall, Laurens van~der Maaten, Laura Gustafson, Maxwell Jones, and
  Aaron~Bryan Adcock.
\newblock Bias amplification in image classification.
\newblock In {\em Workshop on Trustworthy and Socially Responsible Machine
  Learning, NeurIPS 2022}, 2022.

\bibitem{dempster1977maximum}
Arthur~P Dempster, Nan~M Laird, and Donald~B Rubin.
\newblock Maximum likelihood from incomplete data via the em algorithm.
\newblock {\em Journal of the royal statistical society: series B
  (methodological)}, 39(1):1--22, 1977.

\bibitem{wah2011cub}
C~Wah, S~Branson, P~Welinder, P~Perona, and S~Belongie.
\newblock The {Caltech}-{UCSD} {Birds}-200-2011 dataset.
\newblock Technical report, California Institute of Technology, 2011.

\bibitem{zhou2017places}
Bolei Zhou, Agata Lapedriza, Aditya Khosla, Aude Oliva, and Antonio Torralba.
\newblock Places: A 10 million image database for scene recognition.
\newblock {\em IEEE Transactions on Pattern Analysis and Machine Intelligence},
  40(6):1452--1464, 2017.

\bibitem{he2016deep}
Kaiming He, Xiangyu Zhang, Shaoqing Ren, and Jian Sun.
\newblock Deep residual learning for image recognition.
\newblock In {\em Proceedings of the IEEE conference on computer vision and
  pattern recognition}, pages 770--778, 2016.

\bibitem{krizhevsky2017imagenet}
Alex Krizhevsky, Ilya Sutskever, and Geoffrey~E Hinton.
\newblock Imagenet classification with deep convolutional neural networks.
\newblock {\em Communications of the ACM}, 60(6):84--90, 2017.

\bibitem{ye2019unsupervised}
Mang Ye, Xu~Zhang, Pong~C Yuen, and Shih-Fu Chang.
\newblock Unsupervised embedding learning via invariant and spreading instance
  feature.
\newblock In {\em Proceedings of the IEEE/CVF Conference on Computer Vision and
  Pattern Recognition}, pages 6210--6219, 2019.

\bibitem{selvaraju2017grad}
Ramprasaath~R Selvaraju, Michael Cogswell, Abhishek Das, Ramakrishna Vedantam,
  Devi Parikh, and Dhruv Batra.
\newblock Grad-cam: Visual explanations from deep networks via gradient-based
  localization.
\newblock In {\em Proceedings of the IEEE international conference on computer
  vision}, pages 618--626, 2017.

\bibitem{rousseeuw1987silhouettes}
Peter~J. Rousseeuw.
\newblock Silhouettes: A graphical aid to the interpretation and validation of
  cluster analysis.
\newblock {\em Journal of Computational and Applied Mathematics}, 20:53--65,
  1987.

\end{thebibliography}
}

\clearpage
\appendix
\addcontentsline{toc}{section}{Appendix} %
\part{Appendix} %
\parttoc

\section{Additional Dataset Details}
We provide the counts of number of samples in \textit{train}, \textit{val} and \textit{test} splits for our evaluation settings in Table~\ref{tab:splits}. While we use standard splits for CelebA and Waterbirds proposed in prior bias mitigation work~\cite{liu2021jtt}, we propose our own splits for NICO++, which we now describe in more detail.

\noindent\textbf{NICO++ splits.}
To have a sufficient number of samples per-group, we first group concepts (eg. cats, dogs, trains, bikes) to form six super-concepts: \textit{mammals}, \textit{birds}, \textit{plants}, \textit{waterways}, \textit{landways} and \textit{airways}, each occurring in six contexts, giving us a total of $6\times6=36$ groups. We then create settings where each context is spuriously correlated with a unique super-concept (\emph{e.g.} \textit{rocks} (context) and \textit{mammals} (super-class)). This results in 6 bias-aligned and 30 bias-conflicting slices. Table~\ref{tab:nico_high_level} describes the dominant contexts for each super-class, alongwith the base NICO++ classes included in the super-class.

\begin{table}[!h]
\setlength{\tabcolsep}{1.5em}
\begin{tabular}{cccc}
\toprule
    Setting & \textit{train} & \textit{val} & \textit{test} \\
    \midrule
    Waterbirds & 4795 & 1199 & 5794 \\
    CelebA &162770 & 19867 &19962 \\
    \nlo & 9349 & 2349& 1800 \\
    \nmi & 8209 & 2074 & 1800 \\
    \nhi & 7839 & 1979 & 1800\\
    \bottomrule
\end{tabular}
\caption{Number of samples in \textit{train}, \textit{val} and \textit{test} splits of our evaluation settings.\label{tab:splits}}
\end{table}

\begin{table*}[!h]
    \centering
    \begin{tabular}{p{3cm}  p{10cm}  p{2cm}}
    \toprule
       \textbf{Super-concept}  &  \textbf{NICO++ concepts} & \textbf{Dominant context} \\
    \midrule
    mammals  & sheep, wolf, lion, fox, elephant, kangaroo, cat, rabbit, dog, monkey, squirrel, tiger, giraffe, horse, bear, cow & rock\\
    \rowcolor{Gray}
    birds & bird, owl, goose, ostrich & grass\\
    plants & flower, sunflower, cactus & dim lighting\\
    \rowcolor{Gray}
    landways & bicycle, motorcycle, train, bus, scooter, truck, car & autumn\\
    waterways & sailboat, ship, lifeboat & water\\
    \rowcolor{Gray}
    airways & hot air balloon, airplane, helicopter & outdoor\\
    \bottomrule
    \end{tabular}
    \caption{\textbf{NICO++ split details}. We list super-concepts that we use as target labels and their corresponding base concepts from the original NICO++ dataset. Each super-concept co-occurs with six different contexts. In our proposed train and validation splits, each context \emph{dominantly} co-occurs with a unique super-concept 
 (\eg \textit{rocks} and \textit{mammals}) .\label{tab:nico_high_level}}    
\end{table*}

To generate splits, we first randomly select 50 images per each (super-concept, context) pair for creating our evaluation \textit{test} split. For each of \nhi, \nmi and \nlo, we then create a \textit{trainval}  (\textit{train} + \textit{val}) split. 

To create bias-aligned slices from the \textit{trainval} split, we first select a super concept and its corresponding dominant concept (\emph{e.g.} \textit{mammals} and \textit{rocks}) and retrieve all images annotated with both. Next, we select the required number of bias-conflicting samples (where the super concepts occur in \emph{non-dominant} contexts) such that a desired correlation strength of $\beta \in \{75, 90, 95\}$ is ensured for each NICO$++^{\beta}$ setting. Finally, we divide the train-val split uniformly at random in an 80-20 ratio to form the \textit{train} and \textit{val} splits respectively.
Table~\ref{tab:nico_90} shows the resulting train distribution of classes and contexts for NICO$++^{90}$.

\begin{table*}[]
    \centering
    \begin{tabular}{ c c c c c c c}
        \toprule
        \diagbox{Contexts}{Classes} & mammals & birds & plants & airways & waterways & landways \\
        \midrule
        rock & 2552 & 50 & 50 & 50 & 50 & 50 \\
        grass & 24 & 1280 & 24 & 24 & 24 & 24 \\
        dim lighting & 12 & 12 & 616 & 12 & 12 & 12 \\
        outdoor & 16 & 16 & 16 & 879 & 16 & 16 \\
        water & 20 & 20 & 20 & 20 & 1063 & 20 \\
        autumn & 21 & 21 & 21 & 21 & 21 & 1104 \\
        \bottomrule
    \end{tabular}
    \caption{\textbf{\nmi{} split distribution.} Number of samples per each (super-concept, context) pair in the \textit{train} split. }
    \label{tab:nico_90}
\end{table*}

\section{\method: Additional analysis}

\subsection{GradCAM visualizations}

\begin{figure*}[!ht]
    \centering
\includegraphics[width=0.9\linewidth]{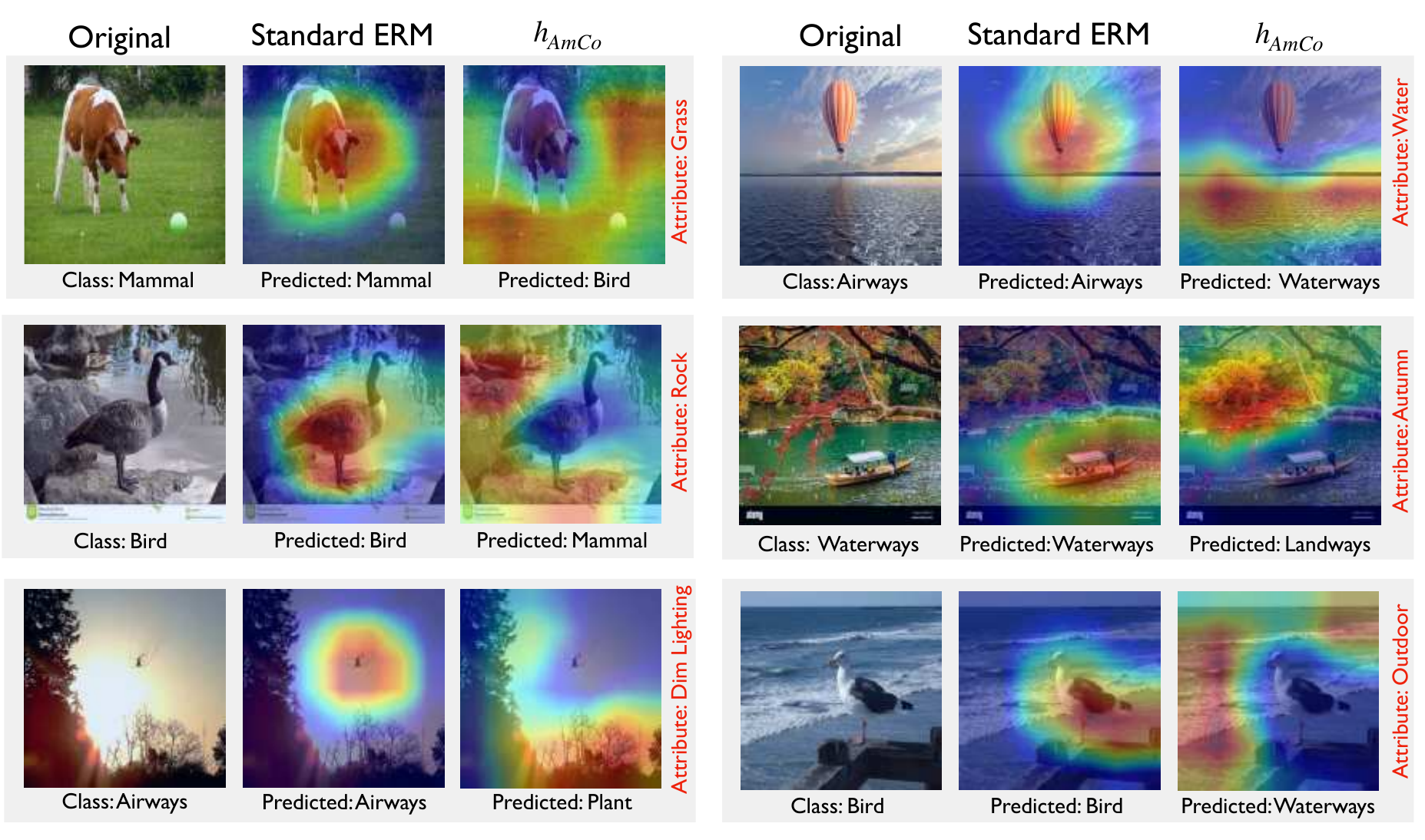}{
\caption{GradCAM \cite{selvaraju2017grad} visualizations of bias-conflicting samples which were correctly classified by the ERM model. We see that the bias-amplified model $h_{\sone}$ makes predictions focusing on the features associated with the spurious attribute, whereas the standard ERM model is able to learn some of the features associated with the target label.}
\label{fig:gradcam_fig}
}
\end{figure*}

We generate GradCAM visualizations \cite{selvaraju2017grad} to investigate the region of interest of the standard ERM model $h_s$ and the bias-amplified model $h_{\sone}$ across the bias-conflicting samples. Figure \ref{fig:gradcam_fig} displays some of the samples belonging to the bias-conflicting slices which were correctly classified using the ERM model. We observe that $h_s$ focuses on features associated with the target label, whereas $h_{\sone}$ focuses on features associated with the spurious attribute. We also observe that $h_{\sone}$ mispredicts the samples as the class that is correlated with the spurious attribute.   

\begin{figure}[!ht]
    \centering
    \includegraphics[width=0.9\linewidth]{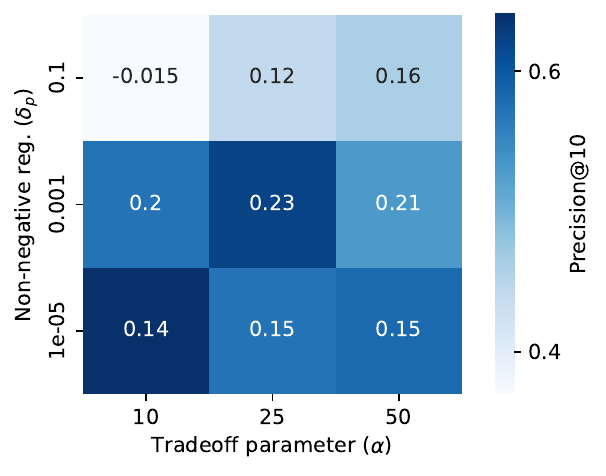}
    \caption{\textbf{Validation of \stwo hyperparameter selection strategy.} We plot the \paten (color, darker is higher) as the hyperparameters in \stwo ($\delta_p$ and $\alpha$) vary along with the values of Silhouette coefficients (inscribed values). We select $\delta_p = 0.001$ and $\alpha=25$ that result in the highest value of Silhouette coefficient.}
    \label{fig:cosi_hparam}
\end{figure}

\subsection{\stwo: Sensitivity to hyperparameters}
For the \nhi{} setting, we use $\alpha=0.25$, $\Sigma_p\cong\textit{full}$ and $\delta_p = 10^{-3}$ as the hyperparameters for the second stage of our approach, \stwo. In Table~\ref{tab:bmm_ablations}, we vary the different hyperparameters used in the second stage of our approach. First, we vary the value of  $\alpha$ - the weight assigned to the \predterm in the log-likelihood (Section 3.2). Next, we change the covariance type of $\Sigma_p$, the covariance of the multivariate normal distribution used to model the \predterm. We find that restricting covariance to \textit{diagonal} 
 drops the \paten to 0.53 from 0.62, while restricting all mixture components to have a \textit{tied} covariance matrix drops the \paten to 0.58. Finally, fitting the mixture model using the logits from our bias-amplified ERM model $h_{\sone}$ to a standard ERM model drops the \paten to 0.36. 

\subsection{\stwo: Hyperparameter tuning and validation}
For tuning the \stwo hyperparameters: $\alpha$, $\delta_p$ and the covariance type of $\Sigma_p$, we recommend using the Silhouette coefficient ~\cite{rousseeuw1987silhouettes}, which is commonly used for unsupervised tuning
of clustering-based algorithms. The Silhouette coefficient~\cite{rousseeuw1987silhouettes} measures how well separated different clusters are in the feature space. In Figure~\ref{fig:cosi_hparam}, we show how the use of Silhouette coefficient~\cite{rousseeuw1987silhouettes} results in selection of hyperparameters achieving good \paten in the \nhi{} setting. Here, we compute the mean of Silhouette coefficients obtained in the embedding spaces of CLIP~\cite{radford2021learning} and model predictions.

\begin{table}[h]
\resizebox*{\linewidth}{!}{
 \begin{tabular}{lc}
\toprule
\textbf{Ablation} & \paten \\
\midrule
$\alpha=25$, $\Sigma_p\cong\textit{full}$, $\delta_p=10^{-3}$ (ours) & \textbf{0.62}\\
\midrule
$\alpha=10$ & 0.57\\
$\alpha=50$ & 0.53 \\
\midrule
$\Sigma_p\cong\textit{diagonal}$ & 0.53\\
$\Sigma_p\cong\textit{tied}$ & 0.58\\
\midrule
$\delta_p=10^{-5}$ & 0.57\\
$\delta_p=10^{-4}$ & \textbf{0.62}\\
$\delta_p=10^{-2}$ & 0.49\\
\midrule
Using standard ERM & 0.36\\
\bottomrule
\end{tabular}
}
{
\caption{\textbf{Ablating \stwo hyperparameters.}  Results on \nhi{}. We report sensitivity of \stwo to hyperparameters by varying one hyperparameter at a time. Changing \cohterm to use logits from a standard ERM trained model results in a huge drop in \paten (last row). %
}
\label{tab:bmm_ablations}.
}
\end{table}

\subsection{Decaying weight decay schedule}
Our current method requires training multiple models with different weight decays for finding the right capacity of model needed. Here, we explore varying the weight decay in a single training run and then picking the model that achieves highest average variation in per-class confidences. Specifically, we decay the weight decay exponentially from $2.0$ to $10^{-3}$ over the course of training. While this simple strategy results in training of far lesser models, we find that this doesn't result in consistent gains in the more difficult NICO++ settings.
\begin{table}[!ht]
    \footnotesize
    \centering
    \setlength{\tabcolsep}{0.5em}
    \begin{tabular}{l c c c c c}
    \toprule
    Identification & Waterbirds & CelebA & NICO$++^{90}$ & NICO$++^{95}$ \\
    \midrule
     \sone  &  0.58 & 0.29 & \textbf{0.41} & \textbf{0.31}\\
      wd schedule   & \textbf{0.63} &  \textbf{0.31}  &0.35 & 0.30\\
    \bottomrule
    \end{tabular}
    \caption{Comparison of methods in terms of \avgap for retrieving bias-conflicting samples across evaluation settings.}
    \label{tab:wd_schedule}
\end{table}

\subsection{Additional qualitative examples}

\begin{figure*}[!ht]
    \centering
  \includegraphics[width=1.0\linewidth]{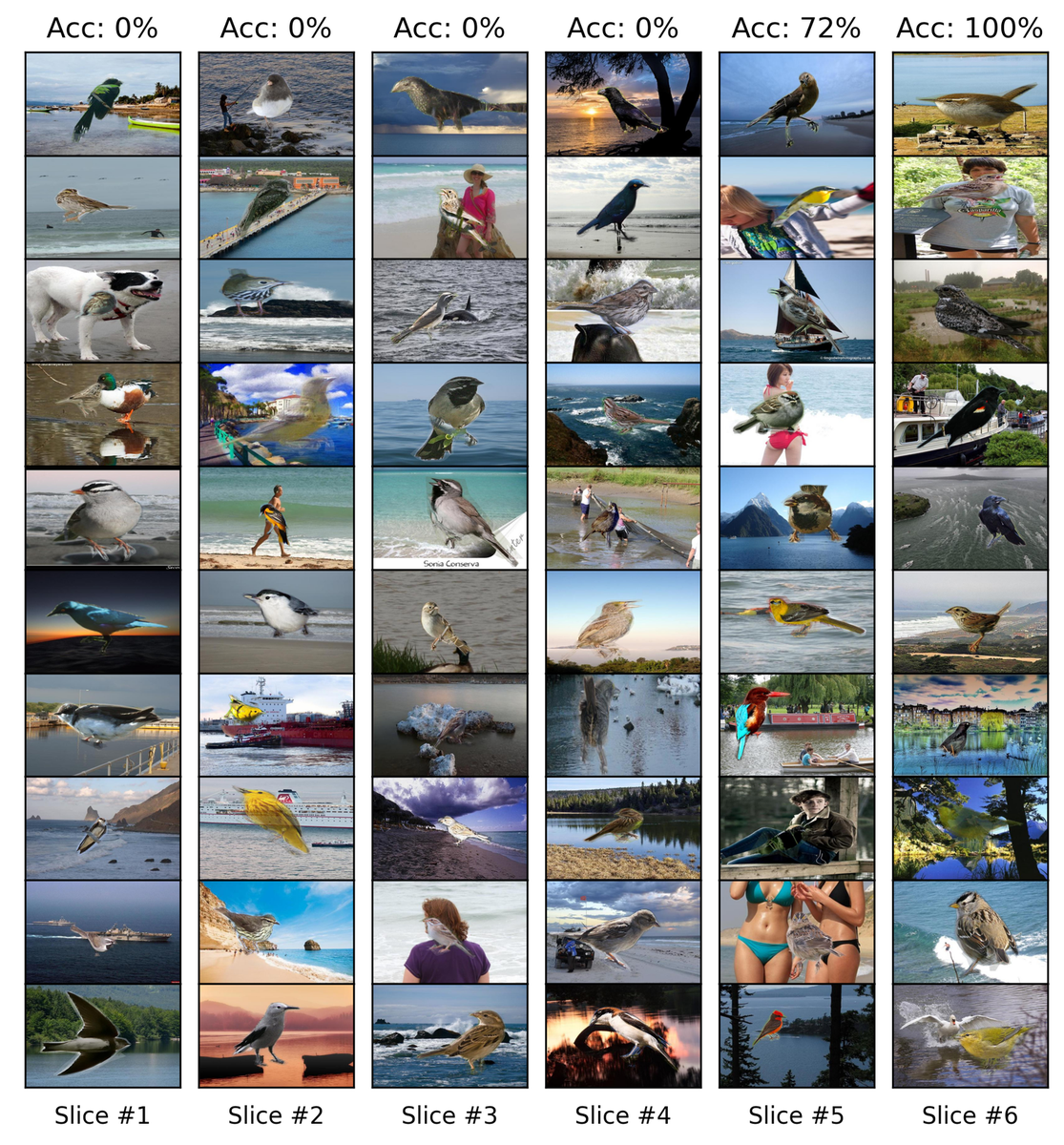}
    \caption{Top-6 slices retrieved by \method for the \textit{landbirds} class from Waterbirds. All these slices predominantly contain the bias-conflicting slices: \textit{landbirds} in \textit{water} backgrounds.}
    \label{fig:qual_landbirds}
\end{figure*}

\begin{figure*}
    \centering
    \includegraphics[width=1.0\linewidth]{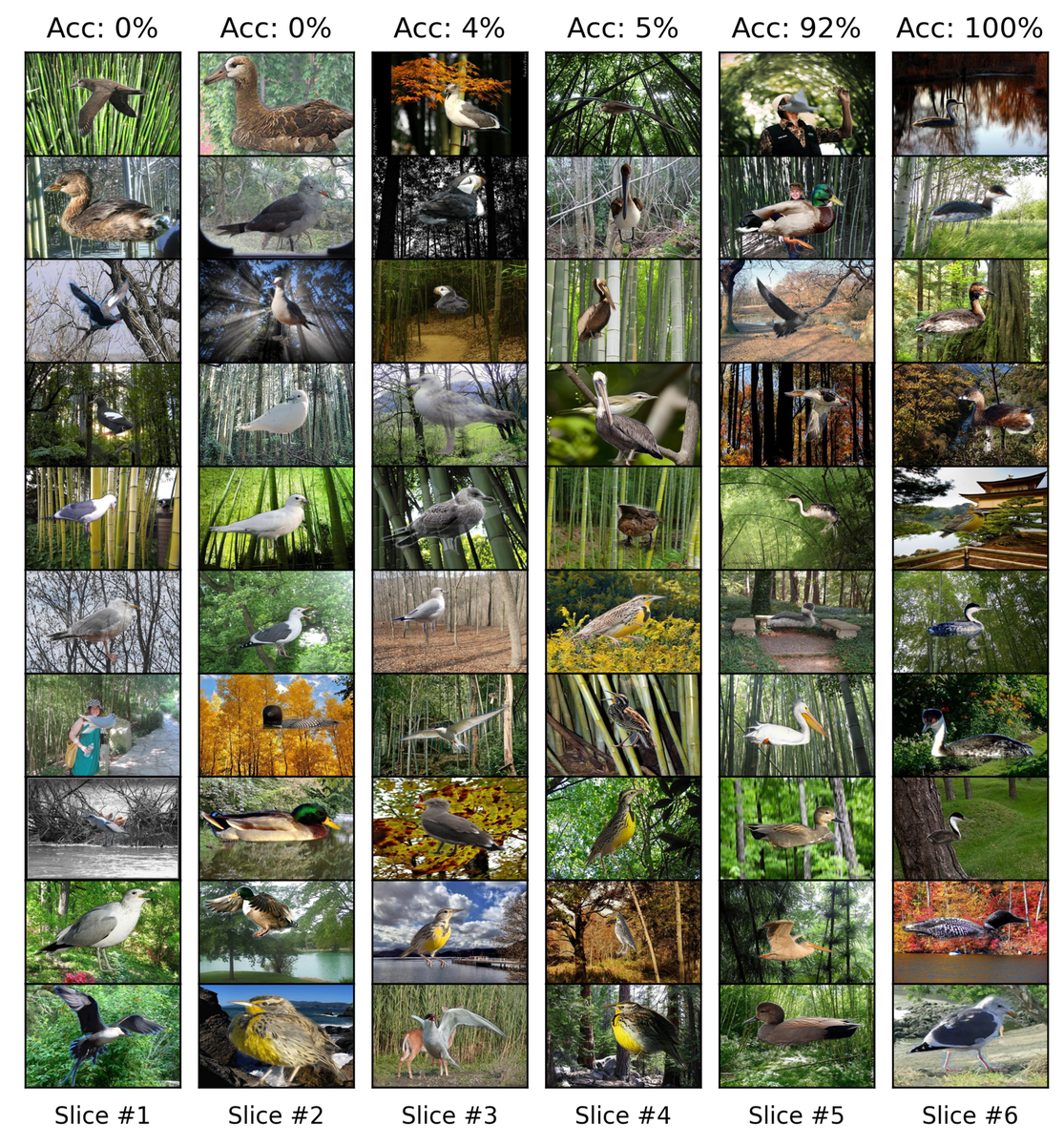}
    \caption{Slices retrieved by \method for the \textit{waterbirds} class from Waterbirds. The samples in these discovered slices are predominantly belong to the bias-conflicting slice of \textit{waterbirds} in \textit{land} backgrounds.}
    \label{fig:qual_waterbirds}
\end{figure*}

\begin{figure*}
    \centering
  \includegraphics[width=1.0\linewidth]{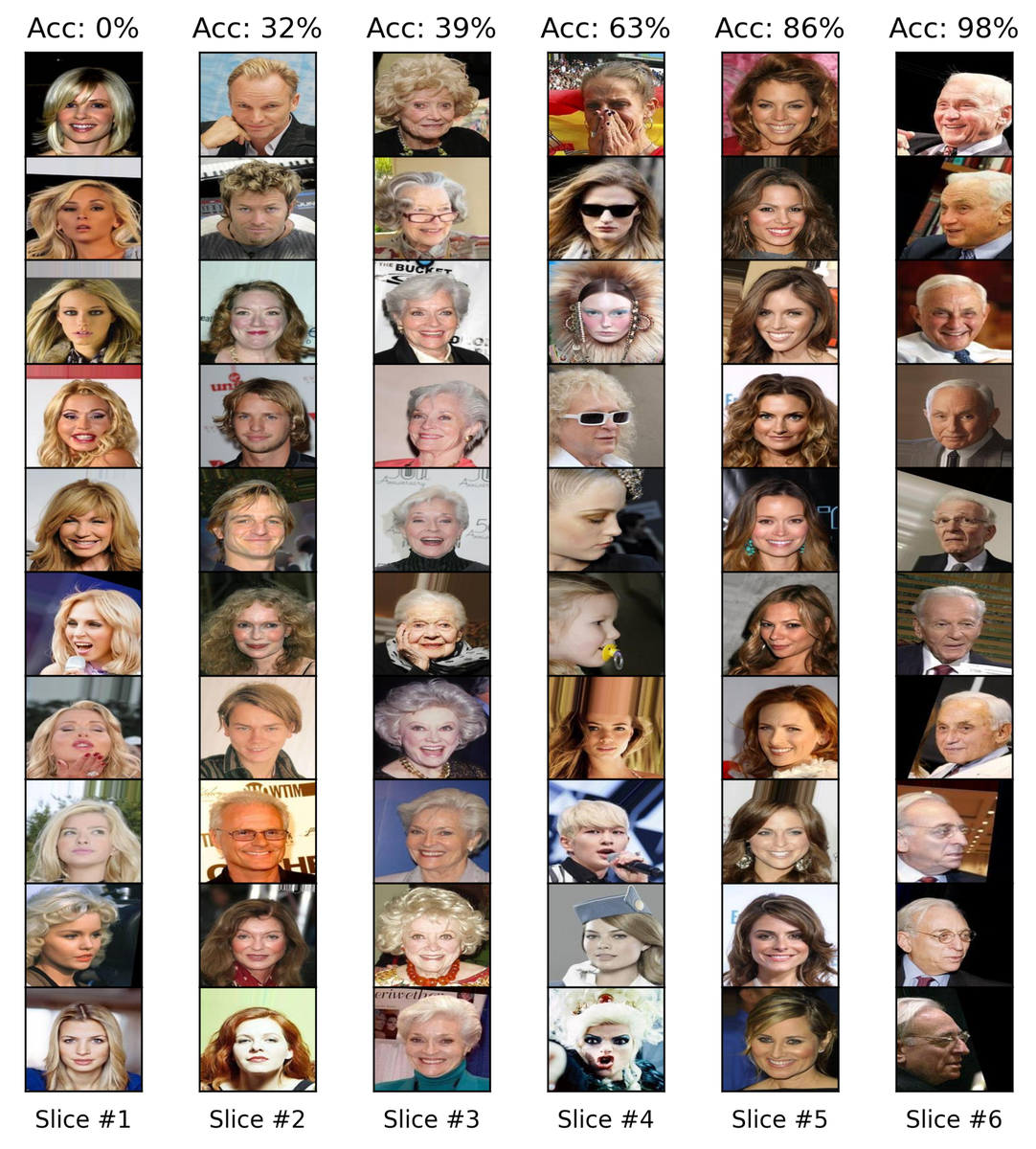}
    \caption{Slices retrieved by \method for the \textit{non-blonde} (dark-haired or gray-haired) class of celebrity faces from CelebA. Note that although this class doesn't have a \textit{bias-conflicting} slice, \method is able to recover coherent slices with degraded performance.}
    \label{fig:qual_non_blonde}
\end{figure*}

\begin{figure*}
    \centering
    \includegraphics[width=1.0\linewidth]{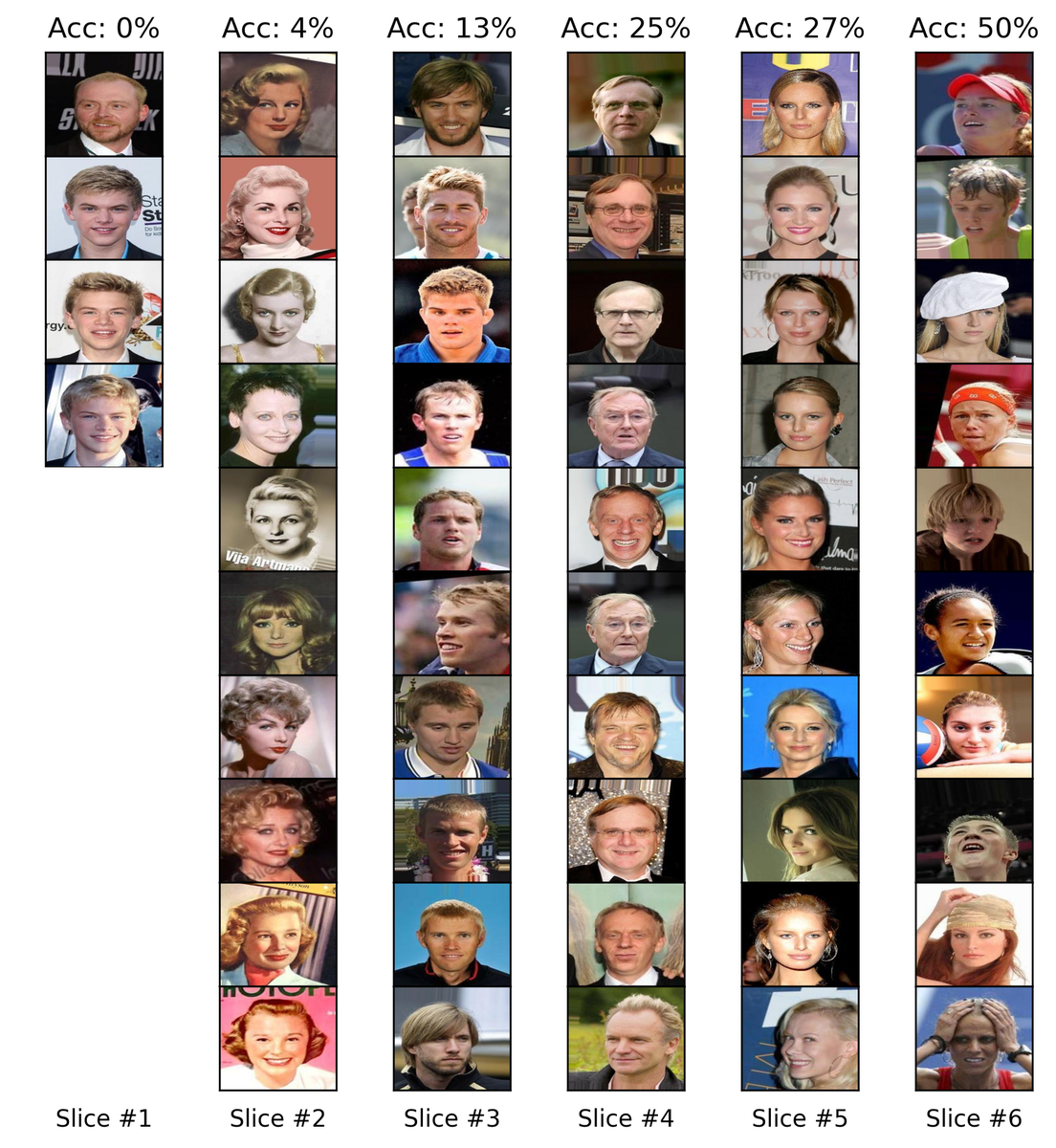}
    \caption{Slices retrieved by \method for the \textit{blonde} class from CelebA. Here, \method recovers multiple slices (Slice \#1, \#3, \#4) corresponding to the \textit{bias-conflicting} slice: \textit{blonde males}.}
    \label{fig:qual_blonde}
\end{figure*}

\begin{figure*}
    \centering
    \includegraphics[width=1.0\linewidth]{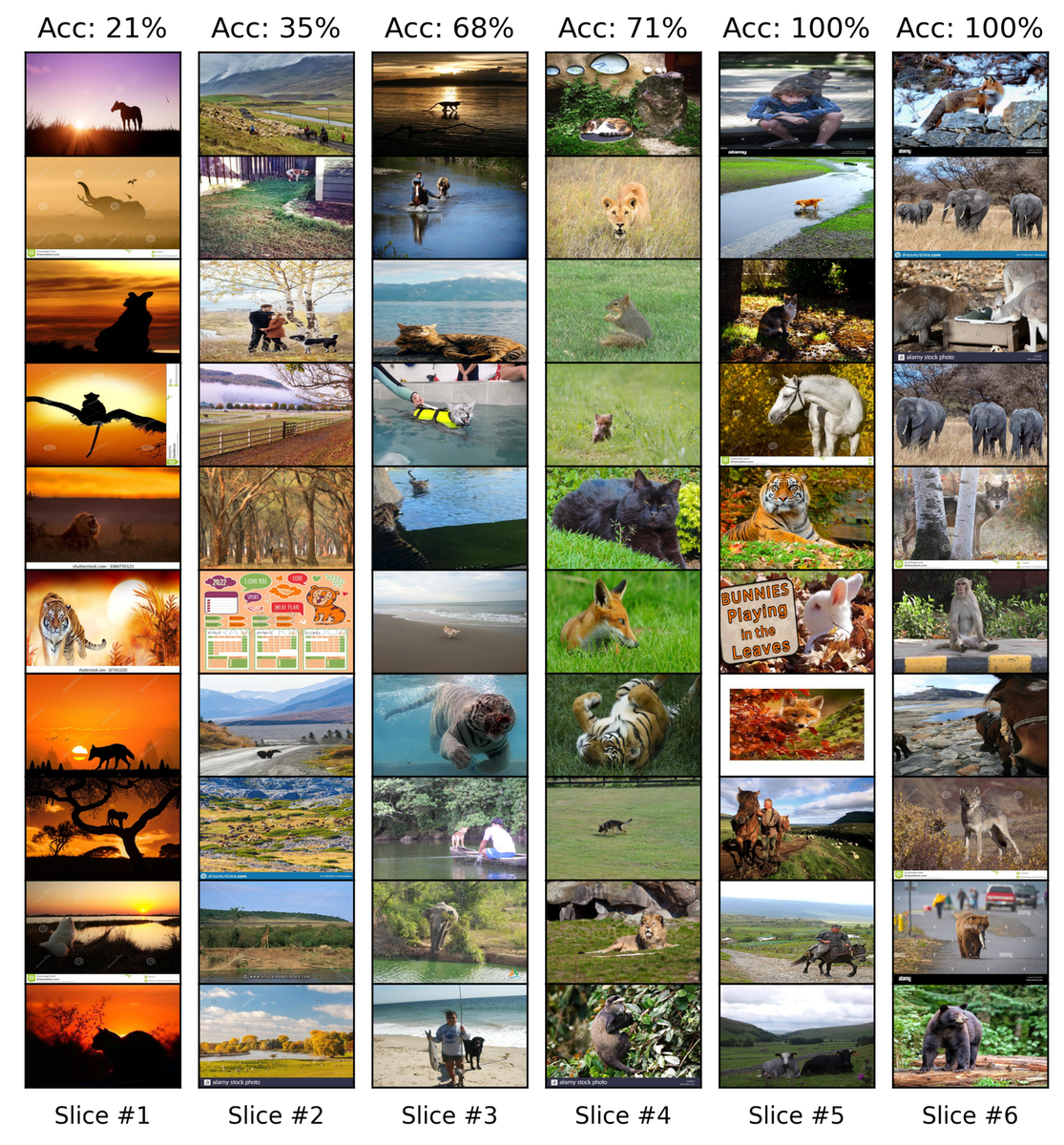}
    \caption{Slices retrieved by \method for the \textit{mammals} class from \nmi{}. Note that the dominant context for \textit{mammals} is \textit{rocks}.}
    \label{fig:qual_mammals}
\end{figure*}

\begin{figure*}
    \centering
    \includegraphics[width=1.0\linewidth]{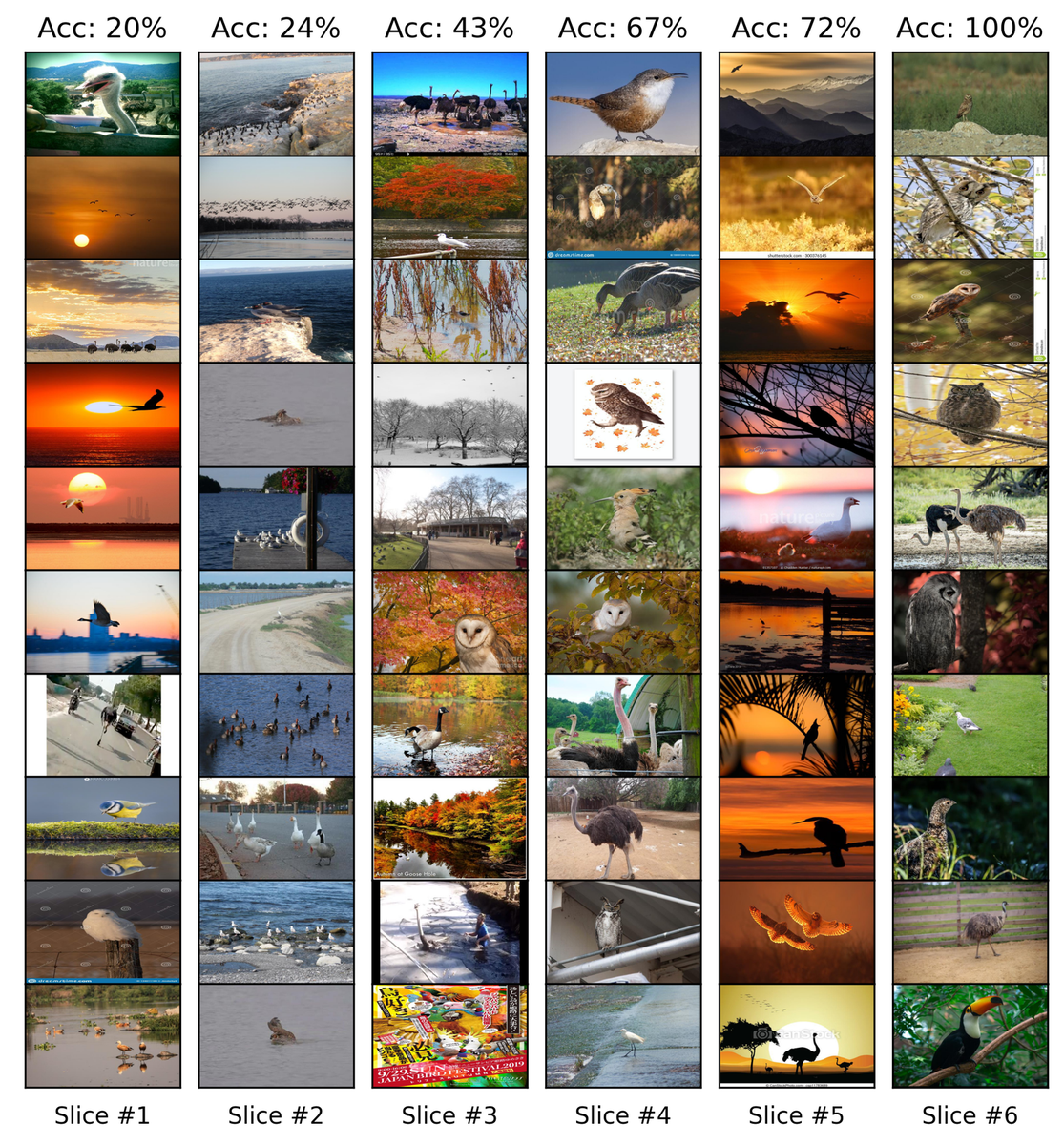}
    \caption{Slices retrieved by \method for the \textit{birds} class from \nmi{}. Note that the dominant context for \textit{birds} is \textit{grass}.}
    \label{fig:qual_birds}
\end{figure*}

\begin{figure*}
    \centering
    \includegraphics[width=1.0\linewidth]{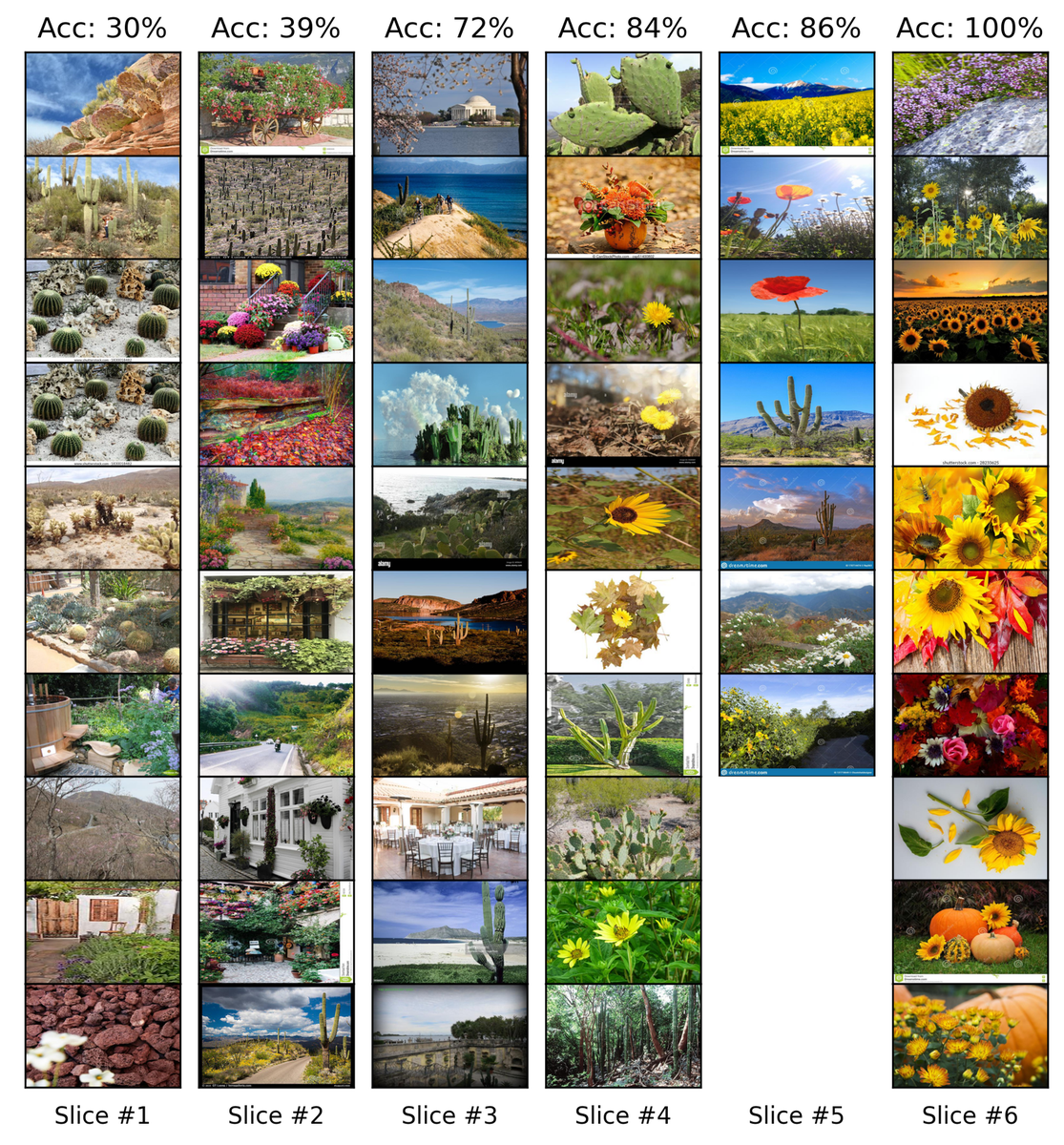}
    \caption{Slices retrieved by \method for the \textit{plants} class from \nmi{}. Note that the dominant context for \textit{plants} is \textit{dim lighting}.}
    \label{fig:qual_plants}
\end{figure*}

\begin{figure*}
    \centering
    \includegraphics[width=1.0\linewidth]{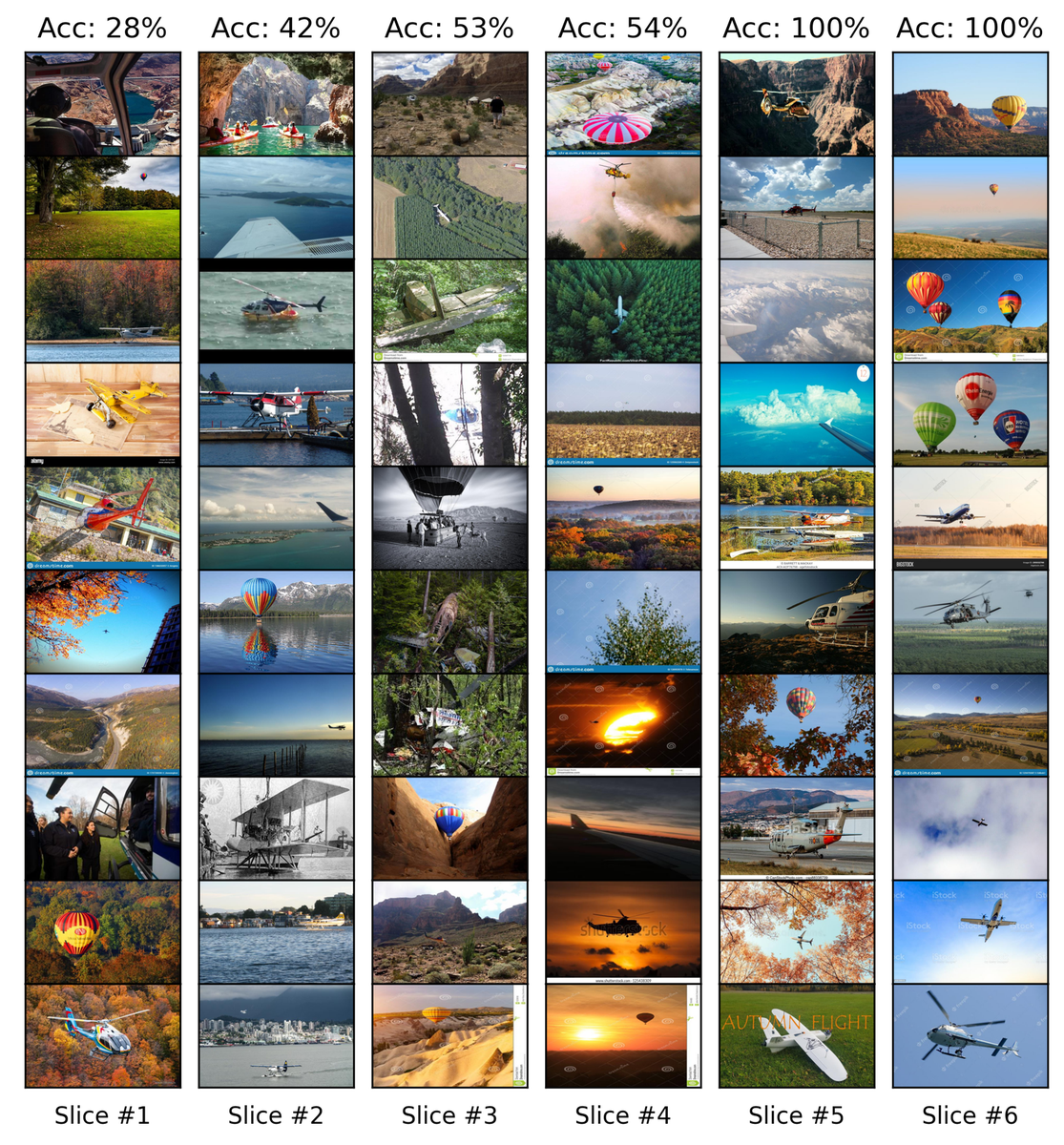}
    \caption{Slices retrieved by \method for the \textit{airways} class from \nmi{}. Note that the dominant context for \textit{airways} is \textit{outdoor}.}
    \label{fig:qual_airways}
\end{figure*}

\begin{figure*}
    \centering
    \includegraphics[width=1.0\linewidth]{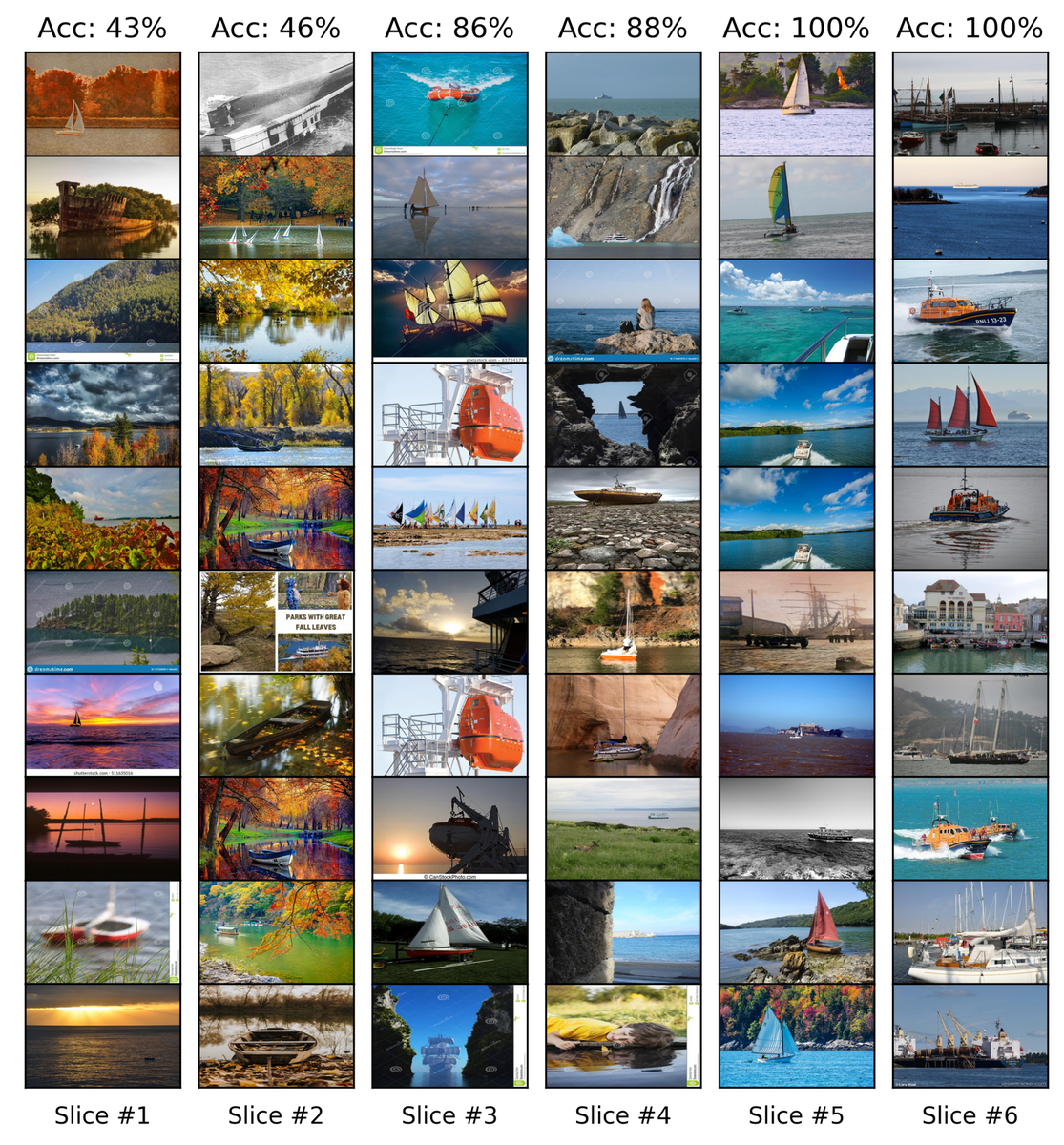}
    \caption{Slices retrieved by \method for the \textit{waterways} class from \nmi{}. Note that the dominant context for \textit{waterways} is \textit{water}.}
    \label{fig:qual_waterways}
\end{figure*}

\begin{figure*}
    \centering
    \includegraphics[width=1.0\linewidth]{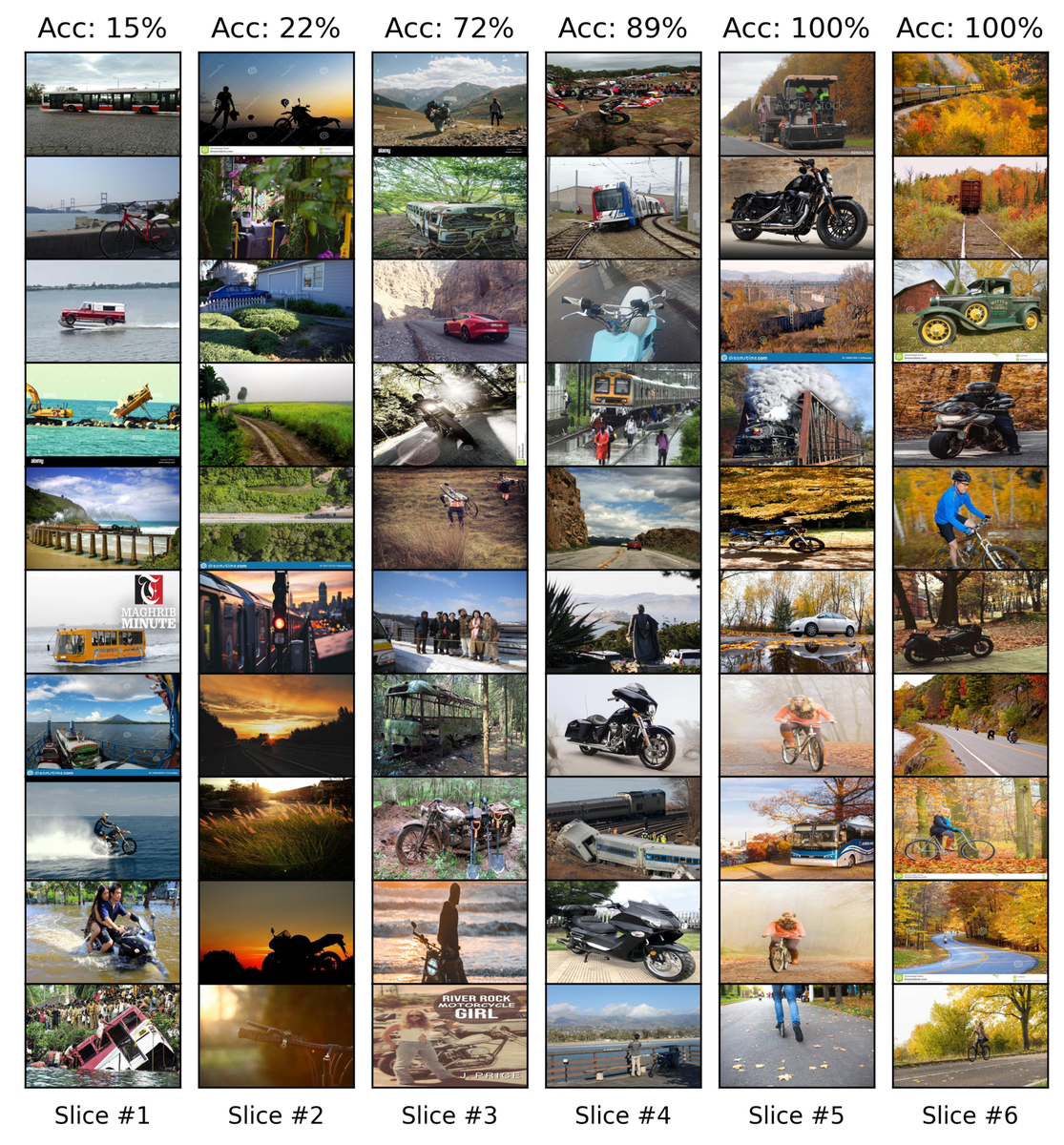}
    \caption{Slices retrieved by \method for the \textit{landways} class from \nmi{}. Note that the dominant context for \textit{landways} is \textit{autumn}.}
    \label{fig:qual_landways}
\end{figure*}

In Figs.~\ref{fig:qual_landbirds}-\ref{fig:qual_landways} we present additional qualitative examples of slices discovered by \method on the CelebA, Waterbirds, and NICO++ settings. We present the top-6 slices after ranking the slices based on model's performance on the slice. We report model's accuracy on each slice at the top.



\section{\method: Additional comparison to prior work}
\subsection{Domino}
\begin{table}[!ht]
   \begin{center} 
   \vspace{-5pt}
   \setlength{\tabcolsep}{2pt}\resizebox{0.99\columnwidth}{!}{
   \begin{tabular}{c c c}
        \toprule
        \textbf{Criterion} & \textbf{FACTS} & \textbf{DOMINO} \\
        \midrule
        Amplification & High $l_2$ reg. & Standard $l_2$ reg. \\
        Prior 
        & Bias amplified logits
        & Cat. assignment of pred. label \\
        Clustering 
        & Per-class, hard assignment 
        & Global, soft assignment \\
        \bottomrule
    \end{tabular}
    }
    \vspace{-10pt}
    \caption{Comparing \method to DOMINO.}  
\label{tab:domino_comparison}
    \end{center}
\end{table}

We summarize the differences with Domino~\cite{eyuboglu2022domino} in Table~\ref{tab:domino_comparison} and describe them below:
\begin{itemize}
\item Our method makes use of a bias-amplified ERM model $h_{\sone}$ instead of the standard ERM model $h_{b}$. This helps increase the separation between the bias-aligned and bias-conflicting samples belonging to a particular class (validated in Section 4.5).

\item Our method makes use of richer prior in the form of logits $h_{\sone}(X)$ produced by the bias-amplified model instead of the categorical assignment of the predicted class label $\hat{Y} = argmax(h_{b} (X))$ using a standard ERM model. 

\item Lastly,~\cite{eyuboglu2022domino} clusters samples across all classes together and separates them by enforcing a soft constraint on the class membership. It does this by jointly modelling its mixture model with class information using the term ${P(\hat{Y} = h_s(x_i)|S^{(j)} = 1)}$. We found  enforcing a hard constraint helps prevent any inter-class contamination of slices.

 \end{itemize}

\subsection{JTT and GroupDRO}
Prior mitigation works~\cite{sagawa2020distributionally,liu2021jtt} use high regularization and low learning rates to achieve good worst group performance. Also,~\cite{sagawa2020distributionally} observes high weight decay to increase the accuracy gap between bias-aligned and bias-conflicting groups. In this work, we exploit this observation for better separating bias-aligned and bias-conflicting groups for the purpose of identifying spurious correlations.

\section{\method: Additional details}

\subsection{{Implementation Details} for \textbf{\stwo}}

In the second stage of our approach, we first initialize the slices using the model's confusion matrix over validation data following~\cite{eyuboglu2022domino}, wherein samples with identical predictions are assigned to the same slices. We fit 36 mixture components per-class. Once the mixture model is fit, we assign each sample to the slice under which the sample achieves the highest density, and rank samples within a slice in order of this density. Finally, we rank slices using model performance and report top-6 slices.

For generating captions and keywords for slices (in Fig. 5) we closely follow~\cite{kim2023explaining}. Specifically, we use an off-the-shelf captioning model, ClipCap~\cite{mokady2021clipcap} for naming all images in a slice. To each slice, we assign a keyword that repeats the most number of times within the captions of the slice.

\end{document}